\documentclass[acmsmall]{acmart}
\setcopyright{acmcopyright}
\copyrightyear{2023}
\acmYear{2023}
\acmDOI{10.1145/1122445.1122456}

\acmJournal{TIST}
\acmVolume{1}
\acmNumber{1}
\acmArticle{1}
\acmMonth{7}

\usepackage{amsfonts}       
\usepackage{nicefrac}       
\usepackage{microtype}      
\usepackage{xcolor}
\usepackage{algorithmic}
\usepackage{algorithm}
\usepackage{color}
\usepackage{xspace}
\usepackage{microtype}
\usepackage{graphicx}
\usepackage{booktabs} 
\usepackage{amsmath}
\usepackage{stackengine}
\usepackage{wrapfig}
\usepackage{multirow}
\usepackage{subcaption}
\floatname{listing}{Listing}

\newcommand{\bm}[1]{\boldsymbol{#1}}

\newcommand{\eg}{\emph{e.g.}}
\newcommand{\ie}{\emph{i.e.}}
\newcommand{\etc}{\emph{etc.}}
\newcommand{\xhdr}[1]{\vspace{1mm} \noindent{{\bf #1.}}}

\newcommand{\set}[1]{\{  #1 \}}

\newcommand{\Ccal}{\mathcal{C}}

\newcommand{\given}{\,|\,}

\newcommand{\nn}{\nonumber}

\newcommand{\hb}{\bm{h}}
\newcommand{\vb}{\bm{v}}
\newcommand{\wb}{\bm{w}}

\newcommand{\Wb}{\bm{W}}

\newcommand{\bb}{\bm{b}}

\newcommand{\nsr}{\textsc{NeuroSeqRet}\xspace}
\newcommand{\nsrf}{\textsc{SelfAttn-}\nsr}
\newcommand{\nsrs}{\textsc{CrossAttn-}\nsr}
\newcommand{\nsrp}{\textsc{Hash-}\nsrs}

\newcommand{\Tcal}{\mathcal{T}}
\newcommand{\Hcal}{\mathcal{H}}
\newcommand{\Mcal}{\mathcal{M}}
\newcommand{\Ucal}{\mathcal{U}}
\newcommand{\Dcal}{\mathcal{D}}
\newcommand{\Bcal}{\mathcal{B}}
\newcommand{\Qr}{\mathcal{Q}}
\newcommand{\Cr}{\mathcal{C}}
\newcommand{\II}{\mathbb{I}}

\newcommand{\sgn}{\mathop{\mathrm{sign}}}
\newcommand{\cp}{\backslash}

\newcommand{\zzc}{\bm{v}^c}
\newcommand{\ehdr}[1]{\vspace{1mm}\noindent\emph{---#1}}
\newcommand{\intensity}{\rho}

\newcommand{\rel}{{+}}
\newcommand{\nrel}{{-}}
\newcommand{\unw}{U}
\newcommand{\pend}{{}}
\newcommand{\kernel}{\kappa}
\newcommand{\para}[1]{\vspace{1mm} \noindent{{#1}}}
\newcommand{\viz}{\emph{viz.}}
\newcommand{\tpprank}{\texttt{Rank}}
\newcommand{\hash}{\bm{\zeta}}
\newcommand{\Wass}{KL}

\newcommand{\bracex}[1]{\left(#1\right)}
\newcommand{\newq}{{q{'}}}

\newcommand{\yb}{\bm{y}}
\newcommand{\ssb}{\bm{s}}
\newcommand{\pb}{\bm{p}}
\newcommand{\kb}{\bm{k}}
\newcommand{\ub}{\bm{u}}
\newcommand{\Ib}{\bm{I}}

\newcommand{\querytt}{\texttt{query}}
\newcommand{\Querytt}{\texttt{Query}}

\newcommand{\keytt}{\texttt{key}}
\newcommand{\Keytt}{\texttt{Key}}

\newcommand{\valuett}{\texttt{value}}
\newcommand{\Valuett}{\texttt{Value}}
\newcommand{\bhbc}{\overline{\bm{h}} ^{(c,q)}}

\newcommand{\eat}[1]{}

\begin{document}

\title{Retrieving Continuous Time Event Sequences using Neural Temporal Point Processes with Learnable Hashing}
\titlenote{This is a substantially refined and expanded version of \citet{neuroseqret} that appeared in the Proc. of AAAI 2022.}

\author{Vinayak Gupta}
\affiliation{
  \institution{University of Washington Seattle}
  \city{Seattle, Washington}
  \country{United States}}
  \email{vinayak@cs.washington.edu}

\author{Srikanta Bedathur}
\affiliation{
  \institution{Indian Institute of Technology Delhi}
  \city{New Delhi}
  \country{India}}
  \email{srikanta@cse.iitd.ac.in}
  
\author{Abir De}
\affiliation{
  \institution{Indian Institute of Technology Bombay}
  \city{Mumbai, Maharashtra}
  \country{India}}
  \email{abir@cse.iitb.ac.in}

\renewcommand{\shortauthors}{Gupta et al.}
\renewcommand{\shorttitle}{CTES Retrieval using Point Processes}

\begin{abstract}
Temporal sequences have become pervasive in various real-world applications such as finance, spatial mobility, health records, \etc\ Consequently, the volume of data generated in the form of continuous time-event sequence(s) or CTES(s) has increased exponentially in the past few years. Thus, a significant fraction of the ongoing research on CTES datasets involves designing models to address downstream tasks such as next-event prediction, long-term forecasting, sequence classification \etc\ The recent developments in predictive modeling using marked temporal point processes (MTPP) have enabled an accurate characterization of several real-world applications involving the CTESs. However, due to the complex nature of these CTES datasets, the task of large-scale retrieval of temporal sequences has been overlooked by the past literature. In detail, by CTES retrieval we mean that for an input query sequence, a retrieval system must return a ranked list of relevant sequences from a large corpus. To tackle this, we propose \nsr, a first-of-its-kind framework designed specifically for end-to-end CTES retrieval. Specifically, \nsr introduces multiple enhancements over standard retrieval frameworks and first applies a trainable unwarping function on the query sequence which makes it comparable with corpus sequences, especially when a relevant query-corpus pair has individually different attributes. Next, it feeds the unwarped query sequence and the corpus sequence into MTPP-guided neural relevance models. We develop four variants of the relevance model for different kinds of applications based on the trade-off between accuracy and efficiency. We also propose an optimization framework to learn binary sequence embeddings from the relevance scores, suitable for the locality-sensitive hashing leading to a significant speedup in returning top-K results for a given query sequence. Our experiments with several datasets show the significant accuracy boost of \nsr beyond several baselines, as well as the efficacy of our hashing mechanism. 
\end{abstract}

\maketitle

\section{Introduction}
Recent developments in marked temporal point processes (MTPP) have dramatically improved the predictive analytics in several real-world applications--- from information diffusion in social networks to healthcare--- by characterizing them with continuous-time event sequences (CTESs)~\cite{Valera2014,rizoiu_hip,wang2017human,daley2007introduction, initiator,du2015dirichlet,srijan,de2016learning,rmtpp,farajtabar2017fake,jing2017neural,proactive}. Though MTPP have shown an incredible prowess in modeling the dynamics of CTES datasets, the task of retrieving \textit{relevant} sequences from a large corpus of CTES data has been overlooked in the past literature. Therefore, in this context, given a query sequence, retrieval of \emph{relevant} CTESs from a corpus of sequences is a challenging problem having a wide variety of search-based applications. For example, in audio or music retrieval, one may like to search sequences having different audio or music signatures; the retrieval of ECG sequences relevant to one pathological query ECG sequence can help in the early detection of cardiac disease; in social network, retrieval of trajectories of information diffusion, relevant to a given trajectory can assist in viral marketing, fake news detection, \etc\ Despite having a rich literature on searching similar time-series~\cite{blondel2021differentiable, gogolou2020data, alaee2020matrix, timegan, cai2019dtwnet, shen2018accelerating, cuturi2017soft, paparrizos2015k, vinayak_thesis}, the problem of designing retrieval models specifically for CTES has largely been unaddressed in the past. Moreover, as shown in our experiments, the existing search methods for time sequences are largely ineffective for a CTES retrieval task, since the underlying characterization of the sequences varies across these two domains. 
 
\subsection{Our Contribution}
In this paper, we first introduce \nsr, a family of supervised retrieval models for continuous-time event sequences, and then develop a trainable locality-sensitive hashing (LSH) based method for efficient retrieval over very large datasets. Specifically, our contributions are as follows:

\xhdr{Query Unwarping} The notion of relevance between two sequences varies across applications. A relevant sequence pair can share very different individual attributes, which can mislead the retrieval model if the sequences are compared as-it-is. In other words, an observed sequence may be a warped transformation of a hidden sequence~\cite{ido,gervini2004self}. To tackle this problem, \nsr\ first applies a trainable unwarping function on the query sequence before the computation of a relevance score. Such an unwarping function is a monotone transformation, which ensures that the chronological order of events across the observed and the unwarped sequences remains the same~\cite{ido}.

\xhdr{Neural Relevance Scoring Model}
In principle, the relevance score between two sequences depends on their latent similarity. We measure such similarity by comparing the generative distribution between the query-corpus sequence pairs. In detail, we feed the unwarped query sequence and the corpus sequence into a neural MTPP-based relevance scoring model which computes the relevance score between the corpus and the unwarped query sequences using two approaches, namely, (i) a Fisher kernel~\cite{fisher} and (ii) a KL-divergence based similarity measure. We highlight that a kernel-based approach offers two key benefits over other distribution similarity measures, \eg, KL divergence or Wasserstein distance: (i) it computes a natural similarity score between query-corpus sequence pairs in terms of the underlying generative distributions; and, (ii) it computes a dot product between the gradients of log-likelihoods of the sequence pairs, which makes it compatible with locality-sensitive hashing for certain design choices and facilitates efficient retrieval. In this context, we provide two variants of \nsr and then propose \nsrp, which allows a nice trade-off between accuracy and efficiency.

\noindent{\underline{\nsrf}}: Here, we use transformer Hawkes process~\cite{thp} which computes the likelihood of corpus sequences independently of the query sequence. Such a design allows the pre-computation of corpus likelihoods, which in turn allows for prior indexing of the corpus sequences before observing the unseen queries. This setup enables us to apply a locality sensitive hashing over the trained embeddings for efficient retrieval.

\noindent{\underline{\nsrs}}: Here, we propose a novel cross attention-based neural MTPP model to compute the sequence likelihoods. Such a cross-attention mechanism renders the likelihood of corpus sequence dependent on the query sequence, making it a more powerful retrieval model. While \nsrs is not directly compatible with such a hashing-based retrieval, it can be employed in a telescopic manner--- where a smaller set of relevant candidates are first retrieved using LSH applied on top of \nsrf, and then re-ranked using \nsrs. Therefore, from the design perspective, these two models provide a trade-off between accuracy and efficiency.

\noindent{\underline{\nsrp}}: This is a novel contribution of this paper and consists of a cross attention-based neural MTPP model that is compatible with a hashing-based retrieval. Unlike \nsrf and \nsrs which rely on a Fisher kernel to compute the similarity between sequences, this approach uses a KL divergence-based distribution similarity approach. Such an approach leads to a significant loss in retrieval performance due to the better modeling capacity of the Fisher kernel, however, it acts as a trade-off between accuracy and efficiency for a cross-attention MTPP model. 

\noindent For all these models, we compute the relevance scores and then learn the unwarping function and the MTPP model by minimizing a pairwise ranking loss. Such a ranking loss is derived from the ground truth relevance labels.

\xhdr{Scalable Retrieval}
Next, we use the predictions made by \nsrf\  and \nsrp\ to develop a novel hashing method that enables efficient sequence retrieval. More specifically, we propose an optimization framework that compresses the learned sequence embeddings into binary hash vectors, while simultaneously limiting the loss due to compression. Then, we use locality-sensitive hashing~\cite{GionisIM1999hash} to bucketize the sequences into hash buckets, so that sequences with similar hash representations share the same bucket. Finally, given a query sequence, we consider computing relevance scores only with the sequences within its bucket. Such a hashing mechanism combined with high-quality sequence embeddings achieves fast sequence retrieval with no significant loss in performance. Finally, our experiments with real-world datasets from different domains show that both variants of \nsr\ outperform several baselines including the methods for continuous-time series retrieval. Moreover, we observe that our hashing method can make a trade-off between retrieval accuracy and efficiency more effectively than baselines based on random hyper-planes as well as exhaustive enumeration.

\subsection{Organization}
The rest of this paper is organized as follows. We review the relevant related work in Section~\ref{sec:relwork} and present a formal problem formulation in Section~\ref{sec:pset}. Section~\ref{sec:model} gives a detailed development of all components in \nsr and Section~\ref{sec:expt} contains in-depth experimental analysis, qualitative, and imputation studies over all datasets before concluding in Section~\ref{sec:conc}.

\section{Related Work}\label{sec:relwork}
Our work is broadly related to the literature of (i) marked temporal point process and (ii) retrieving temporal sequences.

\subsection{Marked Temporal Point Process}
Marked Temporal point processes are central to our work. In recent years, they emerged as a powerful tool to model asynchronous events localized in continuous time~\cite{daley2007introduction, hawkes}, which have a wide variety of applications  \eg, information diffusion, disease modeling, finance, etc. Driven by these motivations, in recent years, there has been a surge of works on MTPP~\cite{rizoiu_hip, rizoiu_sir, du2015dirichlet, farajtabar2017fake, imtpp, reformd, colab}. They predominantly follow two approaches. The first approach which includes the Hawkes process, self-correcting process, \etc, considers fixed parameterization of the temporal point process. Here, different parameterizations characterize the phenomena of interest. In particular, Hawkes process models the self-exciting event arrival process, which is often exhibited by online social networks. However, the fixed parameterization approach often constrains the expressive power of the underlying model, which is often reflected in the sub-optimal predictive performance. The second approach aims to overcome these challenges by modeling MTPP with a deep neural network~\cite{rmtpp, nhp, xiaointaaai, neural_poisson}. For example, \citet{rmtpp} proposed recurrent marked temporal point process (RMTPP)--- an RNN-driven model--- to encapsulate the sequence dynamics and obtain a low dimensional embedding of the event history. This led to further developments which include the Neural Hawkes process that formulates the point process with a continuous-time LSTM~\cite{nhp} and several other neural models of MTPP \eg,~\cite{xiaointaaai, neural_poisson, fullyneural}. However, these approaches are not suitable for sequence retrieval that too for large scale applications. 

\subsection{Retrieving Temporal Event Sequences}
Time series retrieval refers to the process of efficiently retrieving temporal sequences, enabling tasks such as similarity search and pattern recognition for various applications including finance, sensor data analysis, and healthcare. This problem has been addressed in a wide body of research work~\cite{fodo, kasetty, ucrdtw, isax, bakeoff, mass, mueen2016extracting}. However, these techniques are limited to similarity search and standard time series, \ie, ignore the concept of relevance between sequences and the corresponding marks in a CTES. \citet{fodo} introduce the concept of similarity search in time series databases and proposes an efficient algorithm for retrieving similar time series based on similarity measures, while~\citet{kasetty} the need for benchmarks in time series data mining and provides an empirical evaluation of different algorithms for time series retrieval and classification. ~\citet{ucrdtw} present an efficient indexing structure for performing fast similarity searches on time series data using the dynamic time warping (DTW) distance measure.~\cite{isax} proposed indexing techniques for high-dimensional time series data and proposes a method for nearest neighbor searches in high-dimensional spaces. While ~\citet{bakeoff} focuses on classification, it provides a comprehensive review and evaluation of various algorithms and techniques for time series data analysis, including retrieval. Thus, a comprehensive research on retrieving CTES sequences that are relevant to the query is overlooked by the past literature. In detail, the proposed model, \nsr, is designed to handle the marks associated with every event and identify sequences based on their relevancy rather than similarity. 

\section{Preliminaries}
\subsection{Notations and MTPP}
Marked temporal point processes (MTPP) are stochastic processes that capture the generative mechanism of a sequence of discrete events localized in continuous time. Here, an event $e$ is realized using a tuple $(t,x)$, where $t\in\mathbb{R}_+$ and $x\in \mathcal{X}$ are the arrival time and the mark of the event $e$. Then, we use $\Hcal(t)$ to denote a continuous time event sequence (CTES) where each event has arrived until and excluding time $t$, \ie, $\Hcal(t):=\set{e_i=(t_i,x_i)\given t_{i-1}<t_{i}<t}$. Moreover we use $\Tcal(t)$ and $\Mcal(t)$ to denote the sequence of arrival times $\set{t_i\given e_i\in\Hcal(t)}$  and the marks $\set{x_i\given e_i\in\Hcal(t)}$.  Finally, we denote the counting process $N(t)$ as counts of the number of events that happened until and excluding time $t$, encapsulating the generative mechanism of the arrival times.

\xhdr{Generative model for CTES}  
The underlying MTPP model consists of two components -- (i) the dynamics of the arrival times and (ii) the dynamics of the distribution of marks. Most existing works~\cite{rmtpp, sahp,mei_icml,nhp,shelton,thp} model the first component using an intensity function which explicitly models the likelihood of an event in the infinitesimal time window $[t,t+dt)$, \ie,  $\lambda^{\pend}(t )= \text{Pr} (dN(t)=1|\Hcal(t))$. In contrast, we use an intensity-free approach following the proposal by~\citet{intfree}, where we explicitly model the distribution of the arrival time $t$ of the next event $e$. Specifically, we denote the density $\intensity$ of the arrival time and  the distribution $m^{\pend}$ of the mark  of the next event as follows:
\begin{align}\label{eq:qm}
 \intensity(t) dt & = \text{Pr} (e \text{ in } [t,t+dt) \given \Hcal(t)), \\
 m^{\pend}(x)&=\text{Pr} (x\given \Hcal(t))
\end{align}
As discussed by~\citet{intfree}, such an intensity-free MTPP model enjoys several benefits over its intensity-based counterparts, in terms of facilitating efficient training, scalable prediction, computation of expected arrival times, \etc\ Given a sequence of observed events $\Hcal(T)$ collected during the time interval $(0,T]$, the likelihood function is given by:
\begin{equation}
 p (\Hcal(T)) =\textstyle  \prod_{e_i=(t_i,x_i)\in\Hcal(T)}  \intensity(t_i) \times   m^{\pend}(x_i)
\end{equation}

\section{Problem Setup} \label{sec:pset}
Next, we set up our problem of retrieving a ranked list of sequences from a corpus of continuous-time event sequences (CTESs)
which are relevant to a given query CTES. 

\xhdr{Query and corpus sequences, relevance labels} We operate on a large corpus of sequences $\set{\Hcal_c(T_c)\given c\in \Cr}$, where $\Hcal_c(T_c)=\{(t^{(c)} _i, x^{(c)} _i)  \given t^{(c)} _i < T_c \}$. We are given a set of query sequences $\set{\Hcal_q(T_q) \given q\in\Qr}$ with $\Hcal_q(T_q)=\{(t^{(q)} _i, x^{(q)} _i) \given t^{(q)} _i < T_q\}$, as well as a query-specific relevance label for the set of corpus sequences. That is, for a given query sequence $\Hcal_q$, we have: $y(\Hcal_q,\Hcal_c)=+1$  if $\Hcal_c$ is marked as relevant to $\Hcal_q$ and $y(\Hcal_q,\Hcal_c)=-1$ otherwise. 

We define $\Cr_{q\rel} =\{c\in\Cr \given y(\Hcal_q,\Hcal_c)=+1  \}, \text{and, }
\Cr_{q\nrel} =\{c\in\Cr \given y(\Hcal_q,\Hcal_c)=-1  \}$, with $\Cr = \Cr_{q\rel}  \cup \Cr_{q\nrel} $. Finally, we denote $T=\max\{T_q,T_c\given q\in\Qr,c\in\Cr\}$ as the maximum time of the data collection.

\xhdr{Our Goal} We aim to design an efficient CTES retrieval system, which would return a list of sequences from a known corpus of sequences, relevant to a given query sequence $\Hcal_q$. Therefore, we can view a sequence retrieval task as an instance of the ranking problem. Similar to other information retrieval algorithms, a CTES retrieval algorithm first computes the estimated relevance $s(\Hcal_q,\Hcal_c)$ of the corpus sequence $\Hcal_c$ for a given query sequence $\Hcal_q$ and then provides a ranking of $\Cr$ in the decreasing order of their scores.

\section{\nsr Model} \label{sec:model}
In this section, we describe \nsr family of MTPP-based models that we propose for the retrieval of continuous-time event sequences (CTES). We begin with an outline of its two key components.

\subsection{Components of \nsr} \label{subsec:components}
\nsr\  models the relevance scoring function between query and corpus
sequence pairs. However, the relevance of a corpus sequence to the query is latent and varies widely across applications.  To accurately characterize this relevance measure, \nsr\ works in two steps. First, it unwarps the query sequences to make them compatible for comparison with the corpus sequences. Then, it computes the pairwise relevance score between the query and corpus sequences using neural MTPP models.

\xhdr{Unwarping Query Sequences} Direct comparison between a query and a corpus sequence can provide misleading outcomes, since
they also contain their own individual idiosyncratic factors in addition to sharing some common attributes. In fact, a corpus sequence can be highly relevant to the query, despite greatly varying in timescale, initial time, etc. In other words, it may have been generated by applying a warping transformation on a latent sequence~\cite{ido,gervini2004self}. Thus, a direct comparison between a relevant sequence pair may give a poor relevance score. To address this challenge, we first apply a trainable unwarping function~\cite{ido} $U(\cdot)$ on the arrival times of a query sequence, which enhances its compatibility for comparing it with the corpus sequences\footnote{\scriptsize We only apply the unwarping function on the times. Since marks belong to a fixed discrete set, we believe marks are directly comparable.}. More specifically, we define $\unw (\Hcal_q):=\{(\unw(t^{(q)} _i), x^{(q)} _i) \}$. In general, $\unw$  satisfies two properties~\cite{ido,gervini2004self}\eat{ which are}: \emph{unbiasedness}, \ie, having a small value of $\left\|\unw(t )]-t\right\|$ and \emph{monotonicity}, \ie, $ {d\, \unw(t  ) }/{dt} \ge 0$. These properties ensure that the chronological order of the events across both the warped observed sequence and the unwarped sequence remains the same.

Such a sequence transformation learns to capture the similarity between two sequences, even if it is not apparent due to different individual factors, as we shall later in our experiments (Figure~\ref{fig:UU}). 

\xhdr{Computation of Relevance Scores}
Given a query sequence $\Hcal_q$ and a corpus sequence $\Hcal_c$, we compute the relevance score $s(\Hcal_q,\Hcal_c)$ 
using two similarity scores, \eg,  (i) a \emph{model independent} sequence similarity score and (ii) a \emph{model based} sequence similarity score.

\ehdr{Model Independent Similarity Score:} 
Computation of model-independent similarity score between two sequences is widely studied in literature~\cite{xiao2017wasserstein,mueen2016extracting,su2020survey}. They are computed using different distance measures between two sequences, \eg, DTW, Wasserstein distance, \etc, and therefore, can be immediately derived from data without using the underlying MTPP model. In this work, we compute the model-independent similarity score, $\textsc{Sim}_{\unw}(\Hcal_q,\Hcal_c)$, between $\Hcal_q$ and $\Hcal_c$ as follows:
\begin{align}
\textsc{Sim}_{\unw}(\Hcal_q,\Hcal_c)& = -\Delta_{t}(\unw(\Hcal_q),\Hcal_c)-\Delta_{x}(\Hcal_q,\Hcal_c)
\end{align}
where, $\Delta_{t}$ and $\Delta_{x}$ are defined as:
\begin{equation}
\Delta_{t}(\unw(\Hcal_q),\Hcal_c) = \sum_{i=0}^ {H_{\min}} \left| \unw(t^{(q)} _i)- t^{(c)} _i\right| + \hspace{-2mm}\sum_{\substack{t_i \in \Hcal_c\cup\Hcal_q\\ i > |H_{\min}| }} (T -t  _i),
\end{equation}
\begin{equation}
\Delta_{x}(\Hcal_q,\Hcal_c) = \sum_{i=0}^{H_{\min}} \II[x^{(q)} _i \neq x^{(c)}_i] + \big||\Hcal_c|-|\Hcal_q|\big|.
\end{equation}
Here, $H_{\min} = \min\{|\Hcal_q|,|\Hcal_c|\}$, $T=\max\{T_q,T_c\}$ where the events of $\Hcal_q$ and $\Hcal_c$ are gathered until time $T_q$ and $T_c$ respectively; $\Delta_{t}(\unw(\Hcal_q),\Hcal_c)$ is the Wasserstein distance between the unwarped arrival time sequence $\unw(\Hcal_q)$ and the corpus sequence~\cite{xiao2017wasserstein} and, $\Delta_{x}(\Hcal_q,\Hcal_c)$ measures the matching error for the marks, wherein the last term penalizes the marks of last $|\Hcal_c|-|\Hcal_q|$ events of $|\Hcal_c|$.

\ehdr{Model-based Similarity Score using Fisher kernel:}
We hypothesize that the relevance score $s(\Hcal_q,\Hcal_c)$ also depends on a latent similarity that may not be immediately evident from the observed query and corpus sequences even after unwarping. Such similarity can be measured by comparing the generative distributions of the query-corpus sequence pairs. To this end, we first develop an MTPP-based generative model $p_{\theta}(\Hcal)$ parameterized by $\theta$ and then compute a similarity score using the Fisher similarity kernel between the unwarped query and corpus sequence pairs $(\unw(\Hcal_q),\Hcal_c)$~\cite{fisher}. Specifically, we compute the relevance score between the unwarped query sequence $\unw(\Hcal_q)$ and the corpus sequence $\Hcal_c$ as follows:
\begin{align}
\kernel_{p_{\theta}}(\Hcal_q,\Hcal_c) = \vb_{p_{\theta}}(\unw(\Hcal_q)) ^\top \vb_{p_{\theta}}(\Hcal_c),   \label{eq:fisher}
\end{align}
where $\theta$ is the set of trainable parameters; $\vb_{p_{\theta}} (\cdot)$ is given by
\begin{align}
 \vb_{p}(\Hcal) =  \Ib^{-1/2} _{\theta} \nabla_{\theta} \log p_{\theta}(\Hcal)/||\Ib^{-1/2} _{\theta} \nabla_{\theta} \log p_{\theta}(\Hcal)||_2,
\end{align}
$\Ib_{\theta}$ is the Fisher information matrix~\cite{fisher}, \ie,
$\Ib_{\theta} = \mathbb{E}_{\Hcal\sim p_{\theta}(\bullet)}\left[\nabla_{\theta} \log p_{\theta}(\Hcal)\nabla_{\theta} \log p_{\theta}(\Hcal) ^\top\right]$. We would like to highlight that $ \kernel_{p_{\theta}}(\Hcal_q,\Hcal_c)$ in Eq.~\eqref{eq:fisher} is a normalized version of Fisher kernel since $||\vb_{p_{\theta}}(\cdot)||=1$. Thus, $\kernel_{p_{\theta}}(\Hcal_q,\Hcal_c)$ measures the cosine similarity between $ \vb_{p_{\theta}}(\unw(\Hcal_q))$ and $\vb_{p_{\theta}}(\Hcal_c) $. 

Note that, KL divergence or Wasserstein distance could also serve our purpose of computing the latent similarity between the generative distributions. However, we choose the Fisher similarity kernel because of two reasons: (i) it is known to be a natural similarity measure that allows us to use the underlying generative model in a discriminative learning task~\cite{fisher,sewell2011fisher}; and, (ii) unlike KL divergence or other distribution (dis)similarities, it computes the cosine similarity between $ \vb_{p_{\theta}}(\unw(\Hcal_q))$ and $\vb_{p_{\theta}}(\Hcal_c) $, which makes it compatible with locality-sensitive hashing~\cite{charikar2002similarity}. 

\ehdr{Net Relevance Score:}
Finally, we compute the relevance score as:
\begin{align}\label{eq:relevance-score-function}
s_{p,\unw}\left(\Hcal_q,\Hcal_c\right) & =   \kernel_{p}\left(\Hcal_q,\Hcal_c\right) 
  +\gamma \textsc{Sim}_U\left(\Hcal_q,\Hcal_{c}\right)
\end{align}
where $\gamma$ is a hyperparameter. 
 
\subsection{Neural Parameterization of \nsr}
Here, we first present the neural architecture of the unwarping function and then describe the MTPP models used to compute the model-based similarity score in Eq.~\eqref{eq:fisher}. As we describe later, we use two MTPP models with different levels of modeling sophistication, \viz, \nsrf\ and \nsrs. In \nsrf, the likelihood of a corpus sequence is computed independently of the query sequence using self attention-based MTPP model, \eg, Transformer Hawkes Process~\cite{thp}. As a result, we can employ a locality-sensitive hashing-based efficient retrieval based \nsrf. In \nsrs, on the other hand, we propose a more expressive and novel cross-attention MTPP model, where the likelihood of a corpus sequence is dependent on the query sequence. Thus, our models can effectively trade off between accuracy and efficiency.

\xhdr{Neural Architecture of $U(\cdot)$}  As discussed in Section~\ref{subsec:components}, the unwarping function $U(\cdot)$ should be unbiased \ie, a small value of $\left\|\unw(t)-t\right\|$ and monotonic, \ie, $d \unw(t) /dt > 0$. To this end, we model $U(\cdot)\approx U_{\phi}(\cdot)$ using a nonlinear monotone function which is computed using an unconstrained monotone neural network (UMNN)~\cite{umnn},\ie,
\begin{equation}
 U_{\phi}(t) = \int_0 ^t u_{\phi}(\tau) d\tau + \eta,\label{eq:umnn}
\end{equation}
where $\phi$ is the parameter of the underlying neural network $u_{\phi}(\cdot)$, $\eta\in \mathcal{N}(0,\sigma)$ and $u_{\phi}:\mathbb{R}\to\mathbb{R}_+$ is a non-negative non-linear function. Since the underlying monotonicity can be achieved only by enforcing non-negativity of the integrand $u_{\phi}$, UMNN admits an unconstrained, highly expressive parameterization of monotonic functions.  Therefore, any complex unwarping function $U_{\phi}(\cdot)$ can be captured using Eq.~\eqref{eq:umnn}, by integrating a suitable neural model augmented with ReLU$(\cdot)$ in the final layer. In other words, if $u_{\phi}$ is a universal approximator for a positive function, then $U_{\phi}$  can capture any differentiable unwarping function. We impose an additional regularizer $\frac{1}{\sigma^2}\int_0 ^T \left\|u_{\phi}(t) -1\right\|^2 dt $ on our training loss  which ensures that $\left\|\unw(t)-t\right\|$ remains small.

\xhdr{Neural Architecture of MTPP model $p_{\theta}(\cdot)$}
We provide two variants of $p_{\theta}(\cdot)$, which leads to two retrieval models, \viz, \nsrf\ and \nsrs. These two models offer a nice trade-off between accuracy and efficiency.

\para{\nsrf}: Here, we use the Transformer Hawkes process~\cite{thp} which applies a self-attention mechanism to model the underlying generative process. In this model, the gradient of corpus sequences $\vb_{\theta}(\Hcal_c) =\nabla_{\theta} \log p_{\theta}(\Hcal_c)$ are computed independently of the query sequence $\Hcal_q$. Once we train the retrieval model,  $\vb_{\theta}(\Hcal_c)$ can be pre-computed and bucketized before observing the test query. Such a model, together with the Fisher kernel-based cosine similarity scoring model, allows us to apply locality-sensitive hashing for efficient retrieval. 
 
\para{\nsrs}: The above self-attention-based mechanism models a query agnostic likelihood of the corpus sequences. Next, we introduce a \emph{cross attention} based MTPP model which explicitly takes into account the underlying query sequence while modeling the likelihood of the corpus sequence. Specifically, we measure the latent \eat{cross-attention} {relevance score} between $\Hcal_q$ and $\Hcal_c$ \eat{using the likelihood of a corpus sequence} via a query-induced MTPP model built using the cross-attention between the generative process of both the sequences. 

Given a query sequence $\Hcal_q$ and the first $r$ events of the corpus sequence $\Hcal_c$, we parameterize the generative model for $(r+1)$-th event, \ie, $p (e^{(c)}_{r+1} \given  \Hcal({t_r}))$ as $ p_{\theta  \pend}(\cdot)$, where $p_{\theta  \pend} (e^{(c)} _{r+1}) = \intensity _{\theta  \pend} (t^{(c)} _{r+1})\, m^{\pend}_{\theta  \pend} (x^{(c)} _{r+1})$, where $\intensity$ and $m^{\pend}$ are the density and distribution functions for the arrival time and the mark respectively, as described in Eq.~\eqref{eq:qm}. 

\ehdr{Input Layer:} For each event $e_i ^{(q)}$ in the query sequence $\Hcal_q$ and each event $e_j ^{(c)}$ in the first $r$ events in the corpus sequence $\Hcal_c$,  the input layer computes the initial embeddings $\yb^{(q)} _i$ and $\yb^{(c)} _j$ as follows:
\begin{equation}
\yb^{(q)}_i = \wb_{y, x} x^{(q)}_i + \wb_{y, t}\unw(t^{(q)}_i) + \wb_{y, \Delta t} \left(\unw(t^{(q)}_i) - \unw(t^{(q)}_{i-1})\right) + \bb_{y}, \forall i\in[|\Hcal_q|],
\end{equation}
\begin{equation}
\yb^{(c)}_j = \wb_{y, x} x^{(c)}_j + \wb_{y, t}t^{(c)}_j + \wb_{y, \Delta t} \left(t^{(c)}_j - t^{(c)}_{j-1}\right)   + \bb_{y}, \forall j\in[|\Hcal_c(t_{r})|-1]
\end{equation}
where $\wb_{\bullet,\bullet}$ and $\bb_{y}$ are trainable parameters.

\ehdr{Attention layer:}
The second layer models the interaction between all the query events and \emph{the past corpus events}, \ie, $\Hcal_q$ and $\Hcal_c(t_{r})$ using an attention mechanism. Specifically, following the existing attention models~\cite{transformer,sasrec,tisasrec} it first adds a trainable position embedding $\pb$ with $\yb$--- the output from the previous layer. As compared to fixed positional encodings~\cite{transformer}, these learnable encodings demonstrate better performances~\cite{sasrec, tisasrec}. More specifically, we have the updates: $\yb^{(q)}_i \leftarrow \yb^{(q)}_i + \pb_i$ and $ \yb^{(c)}_j \leftarrow \yb^{(c)}_j + \pb_j$. Where, $\pb_\bullet\in \mathbb{R}^D$. Next, we apply  two linear transformations on the vectors  $[\yb^{(q)}_i]_{i\in [|\Hcal_q|]}$ and one linear transformation on  $[\yb^{(c)}_j]_{j\in[r]}$, \ie, $\ssb_j = \Wb^S  \yb^{(c)}_j,   \kb_i = \Wb^K  \yb^{(q)}_i,   \vb_i = \Wb^V  \yb^{(q)}_i$. The state-of-the-art works on attention models~\cite{transformer,thp,sasrec,tisasrec} often refer $\ssb_\bullet$, $\kb_\bullet$ and $\vb_\bullet$ as \querytt,\footnote{\scriptsize The term  \querytt\ in this attention model is different from ``query'' sequence}, \keytt\ and \valuett\ vectors respectively. Similarly, we call the trainable weights $\Wb^S, \Wb^K$ and $\Wb^V$ as the \Querytt, \Keytt\ and \Valuett\ matrices, respectively. Finally, we use the standard attention recipe~\cite{transformer} to compute the final embedding vector $\hb_j ^{(c,q)}$ for the event $e^{(c)} _j$, induced by query $\Hcal_q$. Such a recipe adds the values weighted by the outputs of a softmax function induced by the \querytt\ and \keytt, \ie,
\begin{align}
\hb^{(c,q)} _j = 
 \sum_{i\in [|\Hcal_q|]} \frac{\exp\left( \ssb_j ^\top \kb_i /\sqrt{D} \right)}{\sum_{i'\in [|\Hcal_q|]}\exp\left( \ssb_j ^\top \kb_{i'} /\sqrt{D} \right)} \vb_i, \label{eq:attn}
\end{align}

\ehdr{Output layer:}
Given the vectors $\hb^{(c,q)} _j$ provided by the attention mechanism~\eqref{eq:attn}, we first apply a feed-forward neural network on them to compute  $\bhbc_r$ as follows:
\begin{equation*}
\bhbc _{r} = \sum_{j=1}^r \left[\wb_{\overline{h}}\odot\text{ReLU}(\hb^{(c,q)} _j \odot \wb_{h, f} + \bb_{h, o})  + \bb_{\overline{h}}\right],
\end{equation*}
where $\wb_{\bullet,\bullet}$, $\wb_\bullet$ and $\bb_{v}$. Finally, we use these vectors to compute the probability density of the arrival time of the next event $e^{(c)} _{r+1}$, \ie,  $\intensity_{\theta  \pend}(t_{r+1})$ and the mark distribution $m^{\pend} _{\theta  \pend}(x_{r+1})$. In particular, we realize $\intensity_{\theta  \pend}(t_{r+1})$ using a log-normal distribution of inter-arrival times, \ie,
\begin{align}
 \hspace{-2mm}t^{(c)}_{r+1}- t^{(c)}_r \sim \textsc{LogNormal}\left(\mu_e\left(\bhbc _{r} \right),\sigma^2 _e\left(\bhbc _{r} \right) \right)\nn,
\end{align}
where, $\left[\mu_e\left(\bhbc _{r} \right),\sigma_e\left(\bhbc _{r} \right)\right] = \Wb_{t,q}  \bhbc _{r} \hspace{-2mm}+\bb_{t,q}$. Similarly, we model the mark distribution as,
\begin{align}
\hspace{-2mm} m^{\pend} _{\theta  \pend}(x_{r+1}) =  \dfrac{\exp\left(\wb_{x,m}^\top \bhbc +b_{x,m} \right)}{\sum_{x'\in \mathcal{X}}\exp\left(\wb_{x',m}^\top \bhbc +b_{x',m}\right)}, \label{eq:mark-mtpp-x}
\end{align}
where $\Wb_{\bullet,\bullet}$ are the trainable parameters. Therefore the set of trainable parameters for the underlying MTPP models is $\theta=\set{\Wb^\bullet, \Wb_{\bullet,\bullet}, \wb_\bullet,\wb_{\bullet,\bullet}, \bb_{\bullet},\bb_{\bullet,\bullet}}$.

\xhdr{Tradeoff between Accuracy and Efficiency} Note that\eat{, unlike \nsrf, } the likelihoods of corpus sequences in \nsrs\ depend on the query sequence, making it more expressive than \nsrf.  However, these query-dependent corpus likelihoods cannot be pre-computed before observing the query sequences and thus cannot be directly used for hashing. However, it can be deployed on top of \nsrf, where a smaller set of relevant candidates is first selected using LSH applied on \nsrf\ and then are re-ranked using this \nsrf. Thus, our proposal offers a nice trade-off between accuracy and efficiency.

\subsection{Parameter Estimation}
Given the query sequences $\set{\Hcal_q}$, the corpus sequences $\set{\Hcal_c}$ along with their relevance labels $\set{y(\Hcal_q,\Hcal_c)}$, we seek to find $\theta$ and $\phi$ which ensure that:
\begin{align}
 s_{p_{\theta},\unw_{\phi}}(\Hcal_q,\Hcal_{c_+}) \gg s_{p_{\theta},\unw_{\phi}}(\Hcal_q, \Hcal_{c_-})\, \forall\, c_{\pm} \in \Cr_{q \pm}.
\end{align}
To this aim, we minimize the following pairwise ranking loss~\cite{Joachims2002ranksvm} to estimate the parameters $\theta,\phi$:
\begin{align}
\underset{\theta,\phi} {\text{min}}\sum_{q\in\Qr}\sum_{\substack{c_+\in \Cr_{q\rel},\\ c-\in \Cr_{q\nrel} }}&    \big[s_{p_{\theta},\unw_{\phi}}(\Hcal_q,\Hcal_{c_-})- \nn  s_{p_{\theta},\unw_{\phi}}(\Hcal_q,\Hcal_{c_+}) + \delta \big]_+, 
\end{align}
where, $\delta$ is a tunable margin.

\section{Scalable Retrieval with Hashing}
Once we learn the model parameters $\theta$ and $\phi$, we can rank the set of corpus sequences $\Hcal_c$ in the decreasing order of $s_{p_{\theta},\unw_{\phi}}(\Hcal_{\newq},\Hcal_c)$ for a new query $\Hcal_{\newq}$ and return top$-K$ sequences. Such an approach requires $|\Ccal|$ comparisons per each test query, which can take a huge amount of time for many real-life applications where $|\Cr|$ is high. However, for most practical query sequences, the number of relevant sequences is a very small fraction of the entire corpus of sequences. Therefore, the number of comparisons between query-corpus sequence pairs can be reduced without significantly impacting the retrieval quality by selecting a small number of candidate corpus sequences that are more likely to be relevant to a query sequence. We first briefly describe the traditional random hyperplane-based hashing method and highlight its limitations before presenting our retrieval approach that uses a specific trainable hash code with desirable properties. 

\subsection{Random Hyperplane based Hashing} \label{sec:rh}
Since the relevance between the query and corpus sequence pairs $(\Hcal_q,\Hcal_c)$ is measured using the cosine similarity between the gradient vectors, \ie, $\kernel_{p_{\theta}}(\Hcal_q,\Hcal_c)$, one can use random hyperplane based locality sensitive hashing method for hashing the underlying gradient vectors $\vb_{p_{\theta}} (\Hcal_c)$~\cite{charikar2002similarity}.
Towards this goal, after training \nsrf\, we generate $R$ unit random vectors $\ub_r\in \mathbb{R}^D$ from i.i.d. Normal distributions and then compute a binary hash code, $\hash^c = [\sgn (\ub_1 ^\top \vb_{p_{\theta}}(\Hcal_c)),\dots, \sgn (\ub_R ^\top \vb_{p_{\theta}}(\Hcal_c))] $ for each $c\in \Ccal$. This leads to $2^R$ possible hash buckets $\set{\Bcal}$, where
each corpus sequence is assigned to one hash bucket using the algorithm proposed by~\citet{GionisIM1999hash}.

When we encounter an unseen test query $\Hcal_q$,  we compute the corresponding hash code $\hash^q$, assign it to a bucket $\Bcal$ and finally return \emph{only those sequences} $\Hcal_c$ which were assigned to this bucket $\Bcal$. Thus, for each query, the number of comparisons is reduced from $|\Ccal|$ to $|\Bcal|$, \ie, the number of corpus sequences in the bucket $\Bcal$.
Thus, if the corpus sequences are assigned uniformly across the different buckets, then the expected number of comparisons becomes $|\Ccal|/2^R$, which provides a significant improvement for $R>2$. 

\xhdr{Limitations of Binary Hash Codes} In practice, binary hash codes are not trained from data and consequently, they are not optimized to be uniformly distributed across different hash buckets. Consequently, the assignment of corpus sequences across different buckets may be quite skewed, leading to inefficient sequence retrieval.

\subsection{Trainable Hashing for Retrieval}
Responding to the limitations of the random hyperplane-based hashing method, we propose to learn the hash codes from data, so that they can be optimized for performance.

\xhdr{Computation of a trainable hash code}
We first apply a trainable nonlinear transformation $\Lambda_{\psi}$ with parameter $\psi$ on the gradients $\zzc = \vb_{p_{\theta}}(\Hcal_c)$ and then learn the binary hash vectors $\hash^c = \sgn\bracex{\Lambda_{\psi}\bracex{\zzc}}$  by solving the following optimization, where we use $\tanh\left(\Lambda_{\psi}(\cdot)\right)$ as a smooth approximation of $\sgn\left(\Lambda_{\psi}(\cdot)\right)$.
\begin{align}\label{eq:hash}
 & \min_{\psi} \frac{\eta_1}{|\Cr|}\sum_{c\in\Cr}   \left|\bm{1}^\top \tanh\bracex{\Lambda_{\psi}\bracex{\zzc}}\right| + \frac{\eta_2}{|\Cr|}\sum_{c\in\Cr}   \left\| \left|\tanh\bracex{\Lambda_{\psi}\bracex{\zzc}}\right|-\bm{1}\right\|_1  \\  
 &\qquad + \frac{2\eta_3}{{D \choose 2}}  \cdot \bigg| \sum_{\substack{c\in \Ccal \\ i\neq j\in [D]}}  \tanh\bracex{\Lambda_{\psi}\bracex{\zzc}[i]} \cdot  \tanh\bracex{\Lambda_{\psi} \bracex{\zzc}[j]}\bigg| \nn
\end{align}

\begin{algorithm}[t]
\small
        \caption{Efficient retrieval with hashing}
        \begin{algorithmic}[1]
          \REQUIRE  Trained corpus embeddings
          $\set{\vb^c = \vb_{p_{\theta}} (\Hcal_c)} $ using \nsrf; \\
          new query sequences $\set{\Hcal_{\newq}}$, $K$:
          \# of corpus sequences to return; trained models for \nsrf\ and \nsrs.
          \STATE \textbf{Output:} $\set{L_\newq}$: top-$K$ relevant sequences from $\set{\Hcal_c}$.
          \STATE $\psi\leftarrow\textsc{TrainHashNet}\left(\Lambda_{\psi},[\vb^c]_{c\in\Cr} \right)$
          \STATE \textsc{InitHashBuckets}$(\cdot )$ 
          \FOR{$c\in \Cr$} 
          \STATE $\hash^c\leftarrow \textsc{ComputeHashCode}\left(\zzc;\Lambda_{\psi}\right)$
          \STATE $\Bcal\leftarrow\textsc{AssignBucket}(\hash^c)$
          \ENDFOR
          \FOR{each new query $\Hcal_{\newq}$} 
          \STATE $\vb^{q'}\leftarrow  \nsrf(\Hcal_{\newq})$ 
          \STATE $\hash^{q'}\leftarrow \textsc{ComputeHashCode} (\vb^{q'};  {\Lambda_{\psi}} )$ 
          \STATE  $\Bcal\leftarrow\textsc{AssignBucket}(\hash^{\newq})$ 
            \FOR{$c\in \Bcal$} 
          \STATE $\vb^{\newq} _{\text{cross}}, \vb^{c} _{\text{cross}} \leftarrow \nsrs(\Hcal_{\newq},\Hcal_c)$
          \STATE $s_{p_{\theta},U_{\phi}} (\Hcal_{q'},\Hcal_c) \leftarrow \textsc{Score}(\vb^{\newq} _{\text{cross}}, \vb^{c} _{\text{cross}}, \Hcal_q,\Hcal_c)$
          \ENDFOR
          \STATE $L_{\newq}\leftarrow \textsc{Rank}(\set{s_{p_{\theta},U_{\phi}} (\Hcal_{q'},\Hcal_c)},K)$
        \ENDFOR
        \STATE \textbf{Return}  $\set{L_{\newq}}$
        \end{algorithmic}      \label{alg:key}
\end{algorithm}

Here, $\sum_{i=1}^3 \eta_i=1$. Moreover, different terms in Eq.~\eqref{eq:hash} allow the hash codes, $\hash^c$, to have a set of four desired properties: (i) the first term ensures that the numbers of $+1$ and $-1$ are evenly distributed in the hash vectors $\hash^c=  \tanh\bracex{\Lambda_{\psi}\bracex{\zzc}}$; (ii) the second term encourages the entries of $\hash^c$ to become as close to $\pm 1$ as possible so that $\tanh(\cdot)$ gives an accurate approximation of $\sgn(\cdot)$ ; (iii) the third term ensures that the entries of the hash codes, $\hash^c$, contain independent information and therefore they have no redundant entries.  Trainable hashing has been used in other domains of information retrieval including graph hashing~\cite{qin2020ghashing,roy2020adversarial}, document retrieval~\cite{zhang2010self, salakhutdinov2009semantic}. However, to the best of our knowledge, such an approach has never been proposed for continuous-time sequence retrieval.

\xhdr{Outline of Our Retrieval Method}
We summarize our retrieval procedure in Algorithm~\ref{alg:key}. We are given gradient vectors $\vb^c = \vb_{p_{\theta}}(\Hcal_c)$ obtained by training \nsrf.  Next, we train an additional neural network $\Lambda_{\psi}$ parameterized by $\psi$ (\textsc{TrainHashNet}$()$, line 2), which is used to learn a binary hash vector $\hash^c$ for each sequence $\Hcal_c$. Then these hash codes are used to arrange corpus sequences in different hash buckets (for-loop in lines 4--7) using the algorithm proposed by~\citet{GionisIM1999hash}, so that two sequences $\Hcal_c, \Hcal_{c'}$ lying in the same hash bucket have a very high value of cosine similarity $\cos(\vb^c,\vb^{c'})$. Finally, once a new query $\Hcal_\newq$ comes, we first compute  $\vb^{\newq}$ using the trained \nsrf\ model and then compute the binary hash codes $\hash^{\newq}$ using the trained hash network $\Lambda_\psi$ (lines 9--10). Next, we assign an appropriate bucket $\Bcal$ to it (line 11) and finally compare it with only the corpus sequences in the same bucket, \ie, $\Hcal_c\in\Bcal$ (lines 12--15) using our model. 

\xhdr{Bucket Assignment}
As suggested in~\cite{GionisIM1999hash}, we design multiple hash tables and assign a bucket to the hashcode of a sequence using only a set of bits selected randomly. More specifically, let the number of hash-tables be $M$. Given a query sequence, we calculate its hashcode using the procedure described in Algorithm~\ref{alg:key}, $\hash^{\newq} = \sgn\left(\Lambda_{\psi}(\vb^{\newq})\right)$. The hash code is a $R$ dimension vector with $\hash^{\newq} \in \{-1, 1\}^{R}$ and from this vector, we consider $L$ bits at random positions to determine the bucket to be assigned to the sequence. Here, $\hash^{\newq} $ represents the numbers between $\{0, 2^{L}-1\}$, \ie,  one of the $2^{L}$ different buckets in a hash table. Correspondingly, we assign $\hash^{\newq}$ into a bucket. However, such a procedure is dependent on the specific set of bits --that were selected randomly-- used for deciding the bucket ID. Therefore, we use $M$ hash-tables and repeat the procedure of sampling $L$ bits and bucket assignment for each table. We follow a similar bucket assignment procedure for corpus sequences. As described in Algorithm~\ref{alg:key}, for an incoming query sequence in the test set, we use the above bucket assignment procedure and compute the relevance score for only the corpus sequences within the same buckets. For all our experiments we set $H$ same as the hidden dimension $D$, $M = 10$, and $L=12$.

Recall that we must use \nsrf\ to compute gradient vectors for subsequent hash code generation (in lines 9--10). However, at the last stage for final score computation and ranking,  we can use any variant of \nsr (in line 13), preferably \nsrs, since the corpus sequences have already been indexed by our LSH method.

\section{\nsrp: Cross Attention and Hashing}
The \nsrf is the \textit{hashable} version of \nsr, due to the limitation of the Fisher kernel in modeling two samples from the same distribution. In other words, \nsrs cannot learn end-to-end hashable embeddings as the Fisher kernel has limitations in modeling a \textit{cross-attention} between sequences. In this section, we extend the catalog of the \nsr models and present \nsrp. Unlike the previously defined approaches, \nsrp uses a cross-attention mechanism that can be used to train hash-able vector encodings. In detail, it embeds the corpus sequence and an incoming query sequence using a cross-attention mechanism and later minimizes pairwise ranking loss on the KL divergence between the generative distribution of the query corpus pairs. Moreover, it ensures that the coherence between the order of relevance scores and the ground truth relevance labels is preserved. Here, we explain the working of the cross-attention model that combines the query and corpus sequence. 

\xhdr{Parameter Estimation for \nsrp} Unlike the Fisher kernel, \nsrp learns its' model parameters via a pairwise ranking loss. In detail, given a set of query sequences $\set{\Hcal_q\given q\in\Qr}$, a set of corpus sequences $\set{\Hcal_c\given c \in\Cr}$ along with their relevance labels $\set{y(\Hcal_q,\Hcal)\given q\in \Qr, c\in\Cr}$, we seek to minimize the pairwise ranking loss over the KL divergences between the distributions $p_{\unw(\Hcal_q)}(\bullet\given\Hcal_c)$ and $p_{\unw(\Hcal_q)}(\bullet\given \Hcal_q)$  subject to a set of pairwise constraints ensuring the correct order between relevance scores. we denote the relevance score as $g(\Hcal_q,\Hcal_c)$ between query, corpus pairs $\set{\Hcal_q,\Hcal_c}$. More specifically, we calculate the score via the following:
\begin{equation}
	g(\Hcal_q,\Hcal_c) = \sum_{q\in\Hcal_q} \sum_{c\in \Hcal_c} \Wass \left(p_{\unw(\Hcal_q)}(\bullet\given\Hcal_{c})\left|\right| p_{\unw(\Hcal_q)}(\bullet\given\Hcal_q)\right),
\end{equation}
Thus, for the pair-wise ranking of relevant and non-relevant corpus pairs. Our loss function transforms to the following:
\begin{align}
& \min_{p, \unw} \sum_{q\in\Qr} \sum_{\substack{c_+\in \Cr_{q\rel},\\ c-\in \Cr_{q\nrel}}}  \hspace{-3mm} 
 \Big[\Wass\left(p_{\unw(\Hcal_q)}(\bullet\given\Hcal_{c_+})\left|\right| p_{\unw(\Hcal_q)}(\bullet\given\Hcal_q)\right) - \Wass\left( p_{\unw(\Hcal_q)}(\bullet\given\Hcal_{c_-})\left|\right| p_{\unw(\Hcal_q)}(\bullet\given\Hcal_q) \right)\Big], \nn
  \end{align}
Here, $\Wass(\cdot,\cdot)$ denotes the KL divergence between two distributions; and $p_\bullet(\bullet|\Hcal)$ indicates the probability density of the future events given the previously observed events $\Hcal$.

\xhdr{Utility of Our Training Objective over the Cross Attention Score}
The cross-attention score between $\Hcal_q$ and $\Hcal_c$ is given by $p_{\unw( \Hcal_q)}(\Hcal_c)$--- the likelihood of $\Hcal_c$ induced by $\Hcal_q$. However, it does not allow us to compare the underlying generative distributions which also capture the arrival of future events. On the other hand, the training objective explicitly enumerates the distance between two generative distributions, which leads to more robust training than optimizing only for the relevance score model. Thus, we follow a similar procedure as in \nsr of incorporating model-independent similarity scores, and minimize the following pairwise ranking loss~\cite{Joachims2002ranksvm} to estimate the parameters $\theta,\phi$:
\begin{align}
\underset{\theta,\phi} {\text{min}}\sum_{q\in\Qr}\sum_{\substack{c_+\in \Cr_{q\rel},\\ c-\in \Cr_{q\nrel} }}&    \big[g_{p_{\theta},\unw_{\phi}}(\Hcal_q,\Hcal_{c_-})- \nn  g_{p_{\theta},\unw_{\phi}}(\Hcal_q,\Hcal_{c_+}) + \delta \big]_+, 
\end{align}

\section{Experiments}  \label{sec:expt}
In this section, we provide a comprehensive evaluation of \nsr and our hashing method.

\newcommand{\seq}{\texttt{seq}}
\subsection{Experimental Setup} \label{sec:exp-setup}
\xhdr{Datasets} We evaluate our model and the baselines on seven real-world datasets containing event sequences from various domains. The statistics of all datasets are given in Table~\ref{tab:dset_details}. Across all datasets, $|\Hcal_q| = 5K$ and $|\Hcal_c| = 200K$. We partition the set of queries into 50\% training, 10\% validation, and the rest as test sets. During training, we negatively sample 100 corpus sequences for each query.

\begin{enumerate}
\item \textbf{Audio~\cite{coucke2018snips}:} The dataset contains audio files for spoken commands to a smart-light system and the demographics (age, nationality) of the speaker.  Here, a query corpus sequence pair is relevant if they are from an audio file with a common speaker.

\item \textbf{Sports~\cite{multithumos}:} The dataset contains actions (\eg, run, pass, shoot) taken while playing different sports. We consider the time of action and action class as time and mark of sequence respectively. Here, a query corpus sequence pair is relevant if they are from a common sport.

\item \textbf{Celebrity~\cite{nagrani17}:} In this dataset, we consider the series of frames extracted from youtube videos of multiple celebrities as event sequences where event-time denotes the video-time and the mark is decided upon the coordinates of the frame where the celebrity is located. Here, a query corpus sequence pair is relevant if they are from a video file having a common celebrity.

\item \textbf{Electricity~\cite{refit}:}  This dataset contains the power-consumption records of different devices across smart homes in the UK. We consider the records for each device as a sequence with an event mark as the \textit{normalized} change in the power consumed by the device and the time of recording as event time. Here, a query corpus sequence pair is relevant if they are from a  similar appliance.

\item \textbf{Health~\cite{ecg}:} The dataset contains ECG records for patients suffering from heart-related problems. Since the length of the ECG record for a single patient can be up to 10 million, we generate smaller individual sequences of length 10,000 and consider each such sequence as an independent sequence. The marks and times of events in a sequence are determined using a similar procedure as in Electricity. Here, a query corpus sequence pair is relevant if they are from a common patient.

\item \textbf{Twitter~\cite{nhp}:} In this dataset, we consider the stream of retweets as a sequence with event-time as the time of retweets and mark denotes the social connections of the user -- ordinary user, mildly famous and \textit{influencer}.

\item \textbf{Amazon~\cite{julian}:} In this dataset, we consider the reviews given by users to the products under the category `Movies', `Toys', and `Beauty' on Amazon. We define sequences based on the series of reviews received by a product. More specifically for each product, we consider the time of the written review as the time of the event and the rating as the corresponding mark in the CTES.
\end{enumerate}

For Health, Celebrity, and Electricity datasets, we lack the true ground-truth labeling of relevance between sequences. Therefore, we adopt a heuristic in which, given a dataset $\Dcal$, from each sequence $\seq_q\in\Dcal$ with $q\in[|\Dcal|]$, we first sample a set of sub-sequences $\mathcal{U}_q=\set{\Hcal\subset \texttt{seq}_q}$ with $|\Ucal_q|\sim \text{Unif}\,[200,300]$. For each such collection $\Ucal_q$, we draw exactly one query $\Hcal_q$ uniformly at random from $\Ucal_q$, \ie, $\Hcal_q\sim\Ucal_q$. Then, we define $\Cr=\cup_{q\in[|\Dcal|]}\Ucal_q\cp \Hcal_q$, $\Cr_{q\rel}=\Ucal_q\cp \Hcal_q$ and $\Cr_{q\nrel}=\cup_{c\neq q}\big(\Ucal_c\cp \Hcal_c\big)$.

\xhdr{System Configuration}
All our models were implemented using Pytorch v1.6.0~\footnote{\scriptsize https://pytorch.org/}. We conducted all our experiments on a server running Ubuntu 16.04, CPU: Intel(R) Xeon(R) Gold 6248 2.50GHz, RAM: 377GB, and GPU: NVIDIA Tesla V100. 

\xhdr{Hyperparameters Setup}
We set the hyper-parameters values of \nsr\ as follows: \begin{itemize} \item contribution of model-independent similarity score in Eq.~\eqref{eq:relevance-score-function}, $\gamma = 0.1$; \item margin parameters for parameter estimation, $\delta \in \{0.1, 0.5, 1\}$ and weight for constraint violations, $\lambda \in \{0.1, 0.5, 1\}$; \item weight parameters for hashing objective~\eqref{eq:hash} $\eta_1, \eta_2, \eta_3 \in \{0.1, 0.2, 0.25\}$ and correspondingly $\eta_4 \in \{0.25, 0.4, 0.7\}$. 
\end{itemize}

\begin{table}[t]
\caption{Statistics of the search corpus for all datasets. ${|\Cr_{q\rel}|}/{|\Cr|}$ denotes the ratio of positive corpus sequences to the total sequences sampled for training. The ratio is kept the same for all queries.}
\small
    \centering
    \begin{tabular}{l|ccccccc}
    \toprule
    \textbf{Dataset} & \textbf{Audio} & \textbf{Celebrity} & \textbf{Electricity} & \textbf{Health} & \textbf{Sports} & \textbf{Twitter} & \textbf{Amazon}\\ \hline
    $  {|\Cr_{q\rel}|}/{|\Cr|} $ & 0.25 & 0.23 & 0.20 & 0.28 & 0.30 & 0.23 & 0.16\\
    Total Events & 1M & 50M & 60M & 60M & 430k & 20M & 20M \\
    \# Marks & 5 & 16 & 5 & 5 & 21 & 3 & 5 \\
    \bottomrule
    \end{tabular}
    \label{tab:dset_details}
\end{table}

\begin{table}[t]
\caption{Hyper-parameter values used by fine-tuning the performance on the validation set.}
\centering
    \resizebox{\textwidth}{!}{%
    \begin{tabular}{l|ccccccc}
    \toprule
    \multirow{2}{*}{\textbf{Parameters}} & \multicolumn{7}{c}{\textbf{Datasets}} \\
    \cmidrule(lr){2-8}
    & \textbf{Audio} & \textbf{Celebrity} & \textbf{Electricity} & \textbf{Health} & \textbf{Sports} & \textbf{Twitter} & \textbf{Amazon}\\ \midrule
    $\gamma$ & 0.1 & 0.1 & 0.5 & 0.1 & 0.1 & 0.1 & 0.1\\
    $\delta$ & 0.5 & 0.5 & 0.1 & 0.1 & 0.5 & 0.5 & 0.5\\
    $\{\eta_1, \eta_2, \eta_3\}$ & \{0.4, 0.3, 0.3\} & \{0.4, 0.3, 0.3\} & \{0.4, 0.3, 0.3\} & \{0.5, 0.25, 0.25\} & \{0.5, 0.25, 0.25\} & \{0.4, 0.3, 0.3\} & \{0.5, 0.25, 0.25\} \\
    Batch-size $\mathcal{B}$ & 32 & 32 & 32 & 16 & 16 & 32 & 32\\
    $D$ & 64 & 64 & 48 & 32 & 32 &  64 &  64\\
    \bottomrule
    \end{tabular}
    }
\label{tab:hyperparameters} 
\end{table}

Moreover, the values of training specific parameter values are: \begin{itemize} \item batch-size, $\mathcal{B}$ is selected from $\{16, 32\}$, \ie, for each batch we select $\mathcal{B}$ query sequences and all corresponding corpus sequences; \item hidden-layer dimension for cross-attention model, $D \in \{32, 48, 64\}$; \item number of attention blocks $N_b = 2$; \item number of attention heads $N_h = 1$ and \item UMNN network as a 2 layer feed-forward network with dimension $\{128, 128\}$. \end{itemize} We also add a dropout after each attention layer with probability $p = 0.2$ and an $l_2$ regularizer with over the trainable parameters with the coefficient set to $0.001$. All our parameters are learned using the Adam optimizer. We summarize the details of hyperparameters across different datasets in Table~\ref{tab:hyperparameters}.

\xhdr{Evaluation Metrics}
We evaluate \nsr and the baselines using mean average precision (MAP), NDCG@k, and mean reciprocal rank (MRR). We calculate these metrics as follows:
\begin{equation}
\text{MAP} = \frac{1}{|\Hcal_{q'}|}\sum_{q' \in \Hcal_{q'}} \text{AP}_{q'}, \quad \text{NDCG@k} = \frac{\text{DCG}_k}{\text{IDCG}_k}, \quad \text{MRR} = \frac{1}{|\Hcal_{q'}|} \sum_{q' \in \Hcal_{q'}}\frac{1}{r_{q'}},
\end{equation}
where $\text{AP}_{q'}, \text{DCG}_k, \text{IDCG}_k$, and $r_{q'}$ denote the average precision, discounted cumulative gain at top-$k$ position, ideal discounted cumulative gain (at top-$k$), and the topmost rank of a related corpus sequence respectively. For all our evaluations, we follow a standard evaluation protocol~\cite{sasrec, tisasrec} for our model and all baselines wherein for each query sequence in the test set, we rank all relevant corpus sequences and 1000 randomly sampled non-relevant sequences. All confidence intervals and standard deviations are calculated after 5 independent runs. For all metrics -- MAP, NDCG, and MRR, we report results in terms of percentages with respect to maximum possible value \ie, 1.

\newcommand{\thbest}{\fbox}
\begin{table}[t!]
\centering
\caption{Retrieval quality in terms of Mean Average Precision (MAP in \%) of all the methods across five datasets on the test set. Numbers with the bold font (boxed) indicate the best (best baseline) performer. Results marked \textsuperscript{$\dagger$} are statistically significant (two-sided Fisher's test with $p \le 0.1$) over the best performing state-of-the-art baseline (\tpprank-THP or Sharp). The standard deviation for MASS and UDTW are zero, since they are deterministic retrieval algorithms.}
\resizebox{\textwidth}{!}{%
\begin{tabular}{l|ccccccc}
\toprule
\textbf{Dataset} & \multicolumn{5}{c}{\textbf{Mean Average Precision (MAP) in \%}} \\ \hline 
 & Audio & Celebrity & Electricity & Health & Sports & Twitter & Amazon\\ \hline \hline
MASS~\cite{mass} & 51.1$\pm$0.0 & 58.2$\pm$0.0 & 19.3$\pm$0.0 & 26.4$\pm$0.0 & 54.7$\pm$0.0 & 48.4$\pm$0.0 & 56.1$\pm$0.0\\
UDTW~\cite{ucrdtw} & 50.7$\pm$0.0 & 58.7$\pm$0.0 & 20.3$\pm$0.0 & 28.1$\pm$0.0 & 54.5$\pm$0.0 & 49.5$\pm$0.0 & 56.9$\pm$0.0\\
Sharp~\cite{blondel2021differentiable} & 52.4$\pm$0.2 & 59.8$\pm$0.5 & 22.8$\pm$0.2 & 28.6$\pm$0.2 & \thbest{56.8$\pm$0.3} & 51.8$\pm$0.4 & 57.5$\pm$0.5\\
RMTPP~\cite{rmtpp} & 48.9$\pm$2.3 & 57.6$\pm$1.8 & 18.7$\pm$0.8 & 24.8$\pm$1.2 & 50.3$\pm$2.5 & 47.1$\pm$2.6 & 52.6$\pm$2.1\\
\texttt{Rank}-RMTPP & 52.6$\pm$2.0 & 60.3$\pm$1.7 & 23.4$\pm$0.7 & 29.3$\pm$0.6 & 55.8$\pm$2.1 & 54.4$\pm$1.7 & 57.9$\pm$1.4\\
SAHP~\cite{sahp} & 49.4$\pm$3.2& 57.2$\pm$2.9 & 19.0$\pm$1.8 & 26.0$\pm$2.1 & 53.9$\pm$3.6 & 48.2$\pm$3.4 & 55.2$\pm$2.7\\
\texttt{Rank}-SAHP & 52.9$\pm$1.8& 61.8$\pm$2.3 & 26.5$\pm$1.2 & 31.6$\pm$1.1 & 55.1$\pm$2.3 & 55.3$\pm$1.9 & 57.4$\pm$2.1\\
THP~\cite{thp} & 51.8$\pm$2.3& 60.3$\pm$1.9 & 21.3$\pm$0.9 & 27.9$\pm$0.9 & 54.2$\pm$2.1 & 53.8$\pm$2.1 & 56.7$\pm$1.7\\
\texttt{Rank}-THP & \thbest{54.3$\pm$1.7}& \thbest{63.1$\pm$2.1} & \thbest{29.4$\pm$0.9} & \thbest{33.6$\pm$1.3} &  56.3$\pm$1.9 & \textbf{59.1$\pm$1.1} & \thbest{58.2$\pm$1.8}\\
\hline
\nsrf & 55.8$\pm$1.8 & 64.4$\pm$1.9 & 30.7$\pm$0.7 & 35.9$\pm$0.9 & 57.6$\pm$1.9 & 56.9$\pm$1.3 & 59.4$\pm$1.8\\
\nsrs & 56.2$\pm$2.1 & 65.1$\pm$1.9 & \textbf{32.4$\pm$0.8}\textsuperscript{$\dagger$} & 37.4$\pm$0.9 & \textbf{58.7$\pm$2.1} & 58.4$\pm$1.2 & 61.3$\pm$2.1\\
\nsrp & \textbf{56.8$\pm$2.2}\textsuperscript{$\dagger$} & \textbf{65.2$\pm$2.0}\textsuperscript{$\dagger$} & 32.1$\pm$0.8 & \textbf{37.8$\pm$0.9}\textsuperscript{$\dagger$} & 58.2$\pm$2.0 &  \thbest{58.6$\pm$1.2} & \textbf{61.7$\pm$2.0}\textsuperscript{$\dagger$} \\
\bottomrule
\end{tabular}
}
\label{tab:exp_map}
\vspace{3mm}
\caption{Retrieval quality in terms of NDCG@10 (in \%) of all the methods.}
\resizebox{\textwidth}{!}{%
\begin{tabular}{l|ccccccc}
\toprule
\textbf{Dataset} & \multicolumn{5}{c}{\textbf{NDCG@10 in \%}} \\ \hline 
 & Audio & Celebrity & Electricity & Health & Sports & Twitter & Amazon\\ \hline \hline
MASS~\cite{mass} & 20.7$\pm$0.0 & 38.7$\pm$0.0 & 9.1$\pm$0.0 & 13.6$\pm$0.0 & 22.3$\pm$0.0 & 29.2$\pm$0.0 & 35.9$\pm$0.0\\
UDTW~\cite{ucrdtw} & 21.3$\pm$0.0 & 39.6$\pm$0.0 & 9.7$\pm$0.0 & 14.7$\pm$0.0 & 22.9$\pm$0.0 & 29.5$\pm$0.0 & 36.6$\pm$0.0\\
Sharp~\cite{blondel2021differentiable} & 21.9$\pm$0.2 & 40.6$\pm$0.5 & 11.7$\pm$0.1 & 16.8$\pm$0.1 & 23.7$\pm$0.2 & 30.6$\pm$0.2 & 36.3$\pm$0.4\\
RMTPP~\cite{rmtpp} & 20.1$\pm$1.9 & 39.4$\pm$2.1 & 8.3$\pm$0.8 & 12.3$\pm$0.5 & 19.1$\pm$1.8 & 25.4$\pm$1.8 & 32.3$\pm$1.6\\
\tpprank-RMTPP & 22.4$\pm$1.3 & 41.2$\pm$1.3 & 11.4$\pm$0.4 & 15.5$\pm$0.5 & 23.9$\pm$1.4 & 31.7$\pm$1.2 & 37.4$\pm$1.9\\
SAHP~\cite{sahp} & 20.4$\pm$2.3 & 39.0$\pm$3.1 & 8.7$\pm$1.2 & 13.2$\pm$1.4 & 22.6$\pm$2.5 & 27.1$\pm$2.3 & 34.7$\pm$2.5\\
\tpprank-SAHP & 23.3$\pm$1.4 & 42.1$\pm$1.7 & 13.3$\pm$0.7 & 17.5$\pm$0.9 & 25.4$\pm$1.8 & 32.8$\pm$1.4 & 38.1$\pm$1.2\\
THP~\cite{thp} & 22.1$\pm$1.1 & 40.3$\pm$1.2 & 10.4$\pm$0.6 & 14.4$\pm$0.3 & 22.9$\pm$1.1 & 31.3$\pm$1.8 & 36.3$\pm$1.5\\
\tpprank-THP & \thbest{25.4$\pm$0.9} & \thbest{44.2$\pm$1.0} & \thbest{15.3$\pm$0.4} & \thbest{19.7$\pm$0.4} & \thbest{26.5$\pm$0.9} & \thbest{36.8$\pm$0.9} & \thbest{40.6$\pm$0.8}\\
\hline
\nsrf & 25.9$\pm$1.1 & 45.8$\pm$1.0 & 16.5$\pm$0.5 & 20.4$\pm$0.4 & 27.8$\pm$1.1 & 36.3$\pm$1.0 & 42.5$\pm$0.7\\
\nsrs & \textbf{28.3$\pm$1.1}\textsuperscript{$\dagger$} & 46.9$\pm$1.2 & 18.1$\pm$0.7 & 22.0$\pm$0.4 & 27.9$\pm$1.2 & \textbf{37.9$\pm$1.2} & 43.6$\pm$1.1 \\ 
\nsrp & 27.8$\pm$1.2 & \textbf{47.1$\pm$1.2}\textsuperscript{$\dagger$} & \textbf{18.4$\pm$0.7}\textsuperscript{$\dagger$} & \textbf{22.3$\pm$0.4}\textsuperscript{$\dagger$} & \textbf{28.2$\pm$1.2}\textsuperscript{$\dagger$} & 37.3$\pm$1.1 & \textbf{43.8$\pm$1.0}\textsuperscript{$\dagger$} \\
\bottomrule
\end{tabular}
}
\label{tab:exp_n10}
\end{table}

\begin{table}[t!]
\footnotesize
\centering
\caption{Retrieval quality in terms of mean reciprocal rank (MRR in \%).}
\resizebox{\textwidth}{!}{%
\begin{tabular}{l|ccccccc}
\toprule
\textbf{Dataset} & \multicolumn{5}{c}{\textbf{Mean Reciprocal Rank (MRR)}} \\ \hline 
 & Audio & Celebrity & Electricity & Health & Sports & Twitter & Amazon\\ \hline \hline
MASS~\cite{mass} & 57.3$\pm$0.0 & 63.7$\pm$0.0 & 17.6$\pm$0.0 & 27.2$\pm$0.0 & 61.6$\pm$0.0 & 49.2$\pm$0.0 & 62.6$\pm$0.0\\
UDTW~\cite{ucrdtw} & 58.5$\pm$0.0 & 64.8$\pm$0.0 & 18.7$\pm$0.0 & 29.2$\pm$0.0 & 61.2$\pm$0.0 & 49.5$\pm$0.0 & 62.2$\pm$0.0\\
Sharp~\cite{blondel2021differentiable} & 58.7$\pm$1.7 & 65.4$\pm$2.6 & 19.8$\pm$0.6 & 30.4$\pm$0.7 & 61.1$\pm$2.3 & 50.2$\pm$1.4 & 62.1$\pm$1.8\\
RMTPP~\cite{rmtpp} & 54.2$\pm$3.9 & 64.5$\pm$4.6 & 15.8$\pm$1.2 & 25.2$\pm$1.8 & 56.2$\pm$4.8 & 43.9$\pm$3.2 & 57.2$\pm$4.1\\
\texttt{Rank}-RMTPP & 60.8$\pm$3.7 & 65.6$\pm$4.0 & 23.3$\pm$1.5 & 30.7$\pm$1.7 & 62.7$\pm$3.9 & 61.0$\pm$3.7 & 63.7$\pm$3.9\\
SAHP~\cite{sahp} & 56.9$\pm$4.3 & 64.2$\pm$4.8 & 17.2$\pm$1.3 & 26.7$\pm$2.1 & 59.4$\pm$4.9 & 56.3$\pm$4.2 & 60.4$\pm$4.5\\
\texttt{Rank}-SAHP & 60.3$\pm$2.5 & 66.1$\pm$2.9 & 25.9$\pm$1.2 & 32.6$\pm$1.4 & 62.1$\pm$2.9 & 62.4$\pm$2.7 & 63.0$\pm$2.7\\
THP~\cite{thp} & 58.6$\pm$2.6 & 65.0$\pm$2.9 & 20.1$\pm$1.1 & 28.2$\pm$1.3 & 60.9$\pm$3.3 & 60.1$\pm$2.7 & 61.9$\pm$2.7\\
\texttt{Rank}-THP & \thbest{62.2$\pm$2.8} & \thbest{68.9$\pm$3.0} & \thbest{31.4$\pm$1.4} & \thbest{36.2$\pm$1.7} & \thbest{63.4$\pm$2.9} & \thbest{65.2$\pm$2.7} & \thbest{65.4$\pm$2.8}\\
\hline
\nsrf & 63.6$\pm$2.7 & 69.3$\pm$3.1 & 33.4$\pm$1.6 & 37.9$\pm$1.7 & 64.3$\pm$3.1 & 65.8$\pm$2.7 & 66.4$\pm$2.8 \\
\nsrs & \textbf{64.5$\pm$2.9}\textsuperscript{$\dagger$} & 70.1$\pm$3.3 & 35.2$\pm$1.7 & 40.3$\pm$1.9 & \textbf{66.7$\pm$3.1}\textsuperscript{$\dagger$} & 66.9$\pm$3.0 & \textbf{68.9$\pm$3.1}\textsuperscript{$\dagger$} \\
\nsrp  & 64.1$\pm$3.1 & \textbf{70.4$\pm$3.3} & \textbf{35.6$\pm$1.7}\textsuperscript{$\dagger$} & \textbf{40.5$\pm$1.9}\textsuperscript{$\dagger$} & 65.8$\pm$3.2 & \textbf{67.0$\pm$3.1} & 68.7$\pm$3.2 \\
\bottomrule
\end{tabular}
}
\label{tab:exp_mrr}
\vspace{3mm}
\centering
\caption{Retrieval quality in terms of NDCG@20 (in \%) of all the methods.}
\resizebox{\textwidth}{!}{%
\begin{tabular}{l|ccccccc}
\toprule
\textbf{Dataset} & \multicolumn{5}{c}{\textbf{NDCG@20}} \\ \hline 
 & Audio & Celebrity & Electricity & Health & Sports & Twitter & Amazon\\ \hline \hline
MASS~\cite{mass} & 17.5$\pm$0.0 & 31.4$\pm$0.0 & 8.1$\pm$0.0 & 13.5$\pm$0.0 & 16.3$\pm$0.0 & 24.5$\pm$0.0 & 31.3$\pm$0.0\\
UDTW~\cite{ucrdtw} & 17.9$\pm$0.0 & 32.5$\pm$0.0 & 8.8$\pm$0.0 & 14.4$\pm$0.0 & 16.0$\pm$0.0 & 24.8$\pm$0.0 & 32.0$\pm$0.0\\
Sharp~\cite{blondel2021differentiable} & 18.2$\pm$0.5 & 33.6$\pm$0.7 & 11.9$\pm$0.3 & 15.9$\pm$0.4 & 17.2$\pm$0.7 & 25.2$\pm$0.5 & 31.2$\pm$0.7\\
RMTPP~\cite{rmtpp} & 16.0$\pm$1.1 & 32.2$\pm$1.6 & 7.1$\pm$0.5 & 12.1$\pm$0.7 & 17.1$\pm$1.1 & 21.4$\pm$1.1 & 28.1$\pm$1.4\\
\texttt{Rank}-RMTPP & 20.2$\pm$1.0 & 33.4$\pm$1.4 & 10.5$\pm$0.5 & 15.4$\pm$0.6 & 21.8$\pm$1.2 & 26.2$\pm$1.0 & 32.8$\pm$1.3\\
SAHP~\cite{sahp} & 19.8$\pm$1.4 & 31.7$\pm$2.1 & 7.8$\pm$0.5 & 13.1$\pm$0.9 & 19.2$\pm$1.5 & 22.8$\pm$1.4 & 30.2$\pm$1.9\\
\texttt{Rank}-SAHP & 21.5$\pm$0.9 & 34.1$\pm$1.0 & 12.3$\pm$0.4 & 17.4$\pm$0.6 & 22.9$\pm$1.0 & 27.5$\pm$0.9 & 33.9$\pm$1.0\\
THP~\cite{thp} & 19.7$\pm$0.8 & 32.9$\pm$1.0 & 9.5$\pm$0.4 & 14.4$\pm$0.5 & 20.8$\pm$0.9 & 25.4$\pm$0.8 & 31.8$\pm$0.9\\
\texttt{Rank}-THP & \thbest{21.8$\pm$0.9} & \thbest{37.7$\pm$1.2} & \thbest{14.4$\pm$0.6} & \thbest{19.3$\pm$0.6} & \thbest{23.3$\pm$1.1} & \thbest{32.3$\pm$0.9} & \thbest{36.3$\pm$1.1}\\
\hline
\nsrf & 22.9$\pm$1.3 & 40.3$\pm$1.5 & 16.3$\pm$0.7 & 20.9$\pm$1.0 & 23.8$\pm$1.2 & 32.8$\pm$1.2 & 37.3$\pm$1.3\\\
\nsrs & \textbf{24.2$\pm$1.5}\textsuperscript{$\dagger$} & \textbf{42.0$\pm$1.8}\textsuperscript{$\dagger$} & 17.6$\pm$0.8 & 22.3$\pm$1.0 & 25.7$\pm$1.3 & 33.6$\pm$1.4 & 39.2$\pm$1.4\\\
\nsrp & 23.9$\pm$1.6 & 41.8$\pm$1.8 & \textbf{18.1$\pm$0.9}\textsuperscript{$\dagger$} & \textbf{22.9$\pm$1.1}\textsuperscript{$\dagger$} & \textbf{26.1$\pm$1.3}\textsuperscript{$\dagger$} & \textbf{34.0$\pm$1.5} & \textbf{39.7$\pm$1.4}\textsuperscript{$\dagger$}\\
\bottomrule
\end{tabular}
}
\label{tab:exp_n20}
\end{table}

\xhdr{Baselines} We consider three continuous time-series retrieval models: (i) MASS~\cite{mass}, (ii) UDTW~\cite{ucrdtw} and (iii) Sharp~\cite{blondel2021differentiable}; and, three MTPP models (iv) RMTPP~\cite{rmtpp}, (v) SAHP~\cite{sahp}, and (vi) THP~\cite{thp}. For sequence retrieval with MTPP models, we first train them across all the sequences using maximum likelihood estimation. Then, given a test query $\Hcal_{\newq}$, this MTPP method ranks the corpus sequences $\set{\Hcal_c}$ in decreasing order of their cosine similarity  $\text{CosSim}(\texttt{emb}^{(\newq)},\texttt{emb}^{(c)})$, where $\texttt{emb}^{(\bullet)}$ is the corresponding sequence embedding provided by the underlying MTPP model. In addition, we build supervised ranking models over these approaches, \emph{viz.,} \tpprank-RMTPP, \tpprank-SAHP and \tpprank-THP corresponding to RMTPP, SAHP, and THP. Specifically, \tpprank-MTPP formulates a ranking loss on the query-corpus pairs based on the cosine similarity scores along with the likelihood function to get the final training objective. Therefore, the vanilla MTPP models are used as unsupervised models and the corresponding \tpprank-MTPP models work as supervised models. For all the baselines, we use the official python implementations released by the authors of MASS~\footnote{\scriptsize https://www.cs.unm.edu/$\sim$mueen/MASS.py}, UDTW~\footnote{\scriptsize https://github.com/klon/ucrdtw}, Sharp~\footnote{\scriptsize https://github.com/google-research/soft-dtw-divergences}, RMTPP~\footnote{\scriptsize https://github.com/Networks-Learning/tf\_rmtpp}, SAHP~\footnote{\scriptsize https://github.com/QiangAIResearcher/sahp\_repo}, and THP~\footnote{\scriptsize https://github.com/SimiaoZuo/Transformer-Hawkes-Process} and we thank them for making their codes public. For MASS and UDTW, we report the results using the default parameter values. For Sharp, we tune the hyper-parameter `\textit{gamma}' (for more details see~\cite{blondel2021differentiable}) based on the validation set. In RMTPP, we set the BPTT length to 50, the RNN hidden layer size to 64, and the event embedding size 16. These are the parameter values recommended by the authors. For SAHP and THP, we set the dimension to 128 and the number of heads to 2. The values for all other transformer parameters are similar to the ones we used for the attention part in \nsr.

\xhdr{Evaluation protocol} We partition the set of queries $\Qr$ into 50\% training, 10\% validation, and the rest as test sets. First, we train a retrieval model using the set of training queries. Then, for each test query $\newq$, we use the trained model to obtain a top-$K$ ranked list from the corpus sequences. Next,  we compute the average precision (AP) and discounted cumulative gain (DCG) of each top-$K$ list, based on the ground truth. Finally, we compute the mean average precision (MAP) and NDCG@$K$, by averaging AP and DCG values across all test queries. We set $K \in \{10, 20\}$. 

\subsection{Results on Retrieval Accuracy}
\xhdr{Comparison with Baselines}
First, we compare the retrieval performance of our model against the baselines. Tables~\ref{tab:exp_map}, \ref{tab:exp_n10}, \ref{tab:exp_mrr}, and \ref{tab:exp_n20} summarizes the results in terms of MAP, NDCG@10, MRR, and NDCG@20 respectively. From the results, we make the following observations:
\begin{itemize}
\item Both \nsrs\ and \nsrf\ outperform all the baselines by a substantial margin.
\item \nsrs outperforms \nsrf, since the former has a higher expressive power.
\item The variants of baseline MTPP models trained for sequence retrieval, \ie, \tpprank-RMTPP, \tpprank-SAHP, and \tpprank-THP outperform the vanilla MTPP models.
\item The performances of vanilla MTPPs and the time series retrieval models (MASS, UDTW, and Sharp) are comparable.
\end{itemize}

\begin{table}[t]
    \caption{Ablation study of \nsrs and its variants in terms of MAP (in \%)}
    \centering
    \resizebox{\textwidth}{!}{%
    \begin{tabular}{l|ccccccc}
    \toprule
    \textbf{Variation of} $s_{p_{\theta},U_{\phi}} (\Hcal_q,\Hcal_c)$ & \textbf{Audio} & \textbf{Celebrity} & \textbf{Health} & \textbf{Electricity} & \textbf{Sports} & \textbf{Twitter} & \textbf{Amazon}\\ \hline
$-\Delta_x (\Hcal_q,\Hcal_c) - \Delta_t (U_{\phi}(\Hcal_q),\Hcal_c)$ & 36.1$\pm$0.0 & 43.7$\pm$0.0 & 18.9$\pm$0.0 & 18.9$\pm$0.0 & 41.3$\pm$0.0 & 39.4$\pm$0.0 & 35.7$\pm$0.0 \\
$ \kernel_{p_{\theta}}(\Hcal_q,\Hcal_c)$ & 53.9$\pm$1.9 & 62.5$\pm$1.3 & 33.6$\pm$0.7 & 30.6$\pm$0.9 & 56.3$\pm$2.1 & 56.7$\pm$1.3 & 57.9$\pm$1.8 \\
$\kernel_{p_{\theta}}(\Hcal_q,\Hcal_c) -\gamma \Delta_x (\Hcal_q,\Hcal_c)$ & 54.6$\pm$1.9 & 63.1$\pm$1.4 & 33.7$\pm$0.7 & 30.8$\pm$0.8 & 55.6$\pm$2.0 & 56.3$\pm$1.3 & 58.7$\pm$1.9 \\
$\kernel_{p_{\theta}}(\Hcal_q,\Hcal_c) -\gamma \Delta_t (U_{\phi}(\Hcal_q),\Hcal_c)$ & 55.7$\pm$2.0 & 63.7$\pm$1.8 & 35.9$\pm$0.8 & 31.3$\pm$0.8 & 58.1$\pm$2.0 & 57.4$\pm$1.6 & 59.1$\pm$2.0 \\
\nsrs Without $U_{\phi}(\cdot)$ & 55.2$\pm$2.2 & 62.9$\pm$2.0 & 34.3$\pm$0.9 & 29.7$\pm$1.3 & 56.2$\pm$2.3 & 56.8$\pm$1.4 & 57.3$\pm$1.9 \\
    \nsrs & \textbf{56.2$\pm$2.1} & \textbf{65.1$\pm$1.9} & \textbf{37.4$\pm$0.9} & \textbf{32.4$\pm$0.8} & \textbf{58.7$\pm$2.1} & \textbf{58.4$\pm$1.2} & \textbf{61.3$\pm$2.1}\\
    \bottomrule
    \end{tabular}
    }
    \label{tab:model_ablation}
\end{table}

\xhdr{Ablation Study}
Next, we compare the retrieval performance across four model variants: 
\begin{itemize}
\item our model with only model-independent score \ie, $s_{p_{\theta},U_{\phi}} (\Hcal_q,\Hcal_c) = -\Delta_x (\Hcal_q,\Hcal_c) - \Delta_t (U_{\phi}(\Hcal_q),\Hcal_c)$;
 \label{var:withoutmodel}
\item our model with only model-dependent score, \ie, $s_{p_{\theta},U_{\phi}}(\Hcal_q,\Hcal_c) = \kernel_{p_{\theta}}(\Hcal_q,\Hcal_c)$;
 \label{var:withmodel}
\item our model without any model independent time similarity \ie, $s_{p_{\theta},U_{\phi}}(\Hcal_q,\Hcal_c) = \kernel_{p_{\theta}}(\Hcal_q,\Hcal_c) -\gamma \Delta_x (\Hcal_q,\Hcal_c)$;
 \label{var:withouttime}
\item our model without any model independent mark similarity \ie, $s_{p_{\theta},U_{\phi}}(\Hcal_q,\Hcal_c) = \kernel_{p_{\theta}}(\Hcal_q,\Hcal_c) -\gamma \Delta_t (U_{\phi}(\Hcal_q),\Hcal_c)$;
 \label{var:withoutmark}
\item  our model without unwarping function $U_{\phi}(\cdot)$;
 \label{var:withoutU}
 and
\item the complete design of our model.
\label{var:our}
\end{itemize}

\begin{figure}[t]
 \centering
\begin{subfigure}{0.45\columnwidth}
  \centering
  \includegraphics[height=3cm]{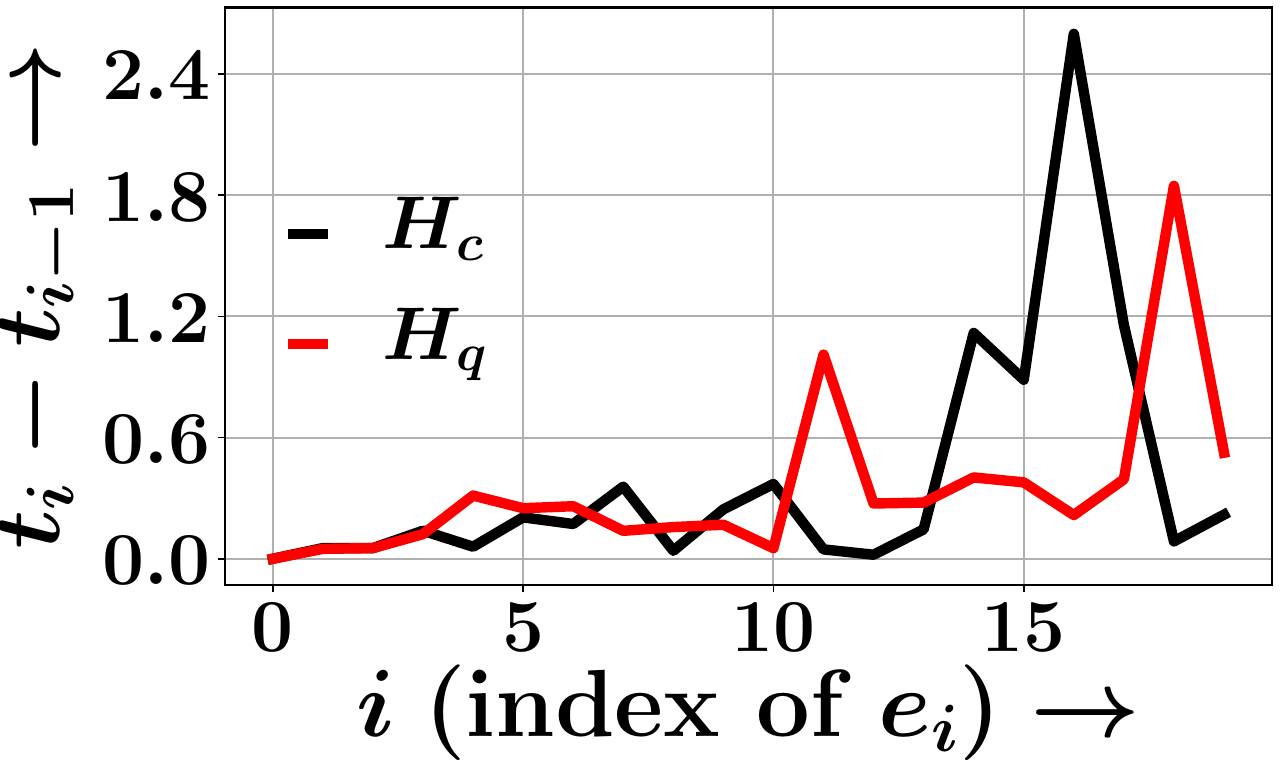}
  \caption{$\Hcal_q, \Hcal_c$}
\end{subfigure}
\hspace{2mm}
\begin{subfigure}{0.45\columnwidth}
  \centering
  \includegraphics[height=3cm]{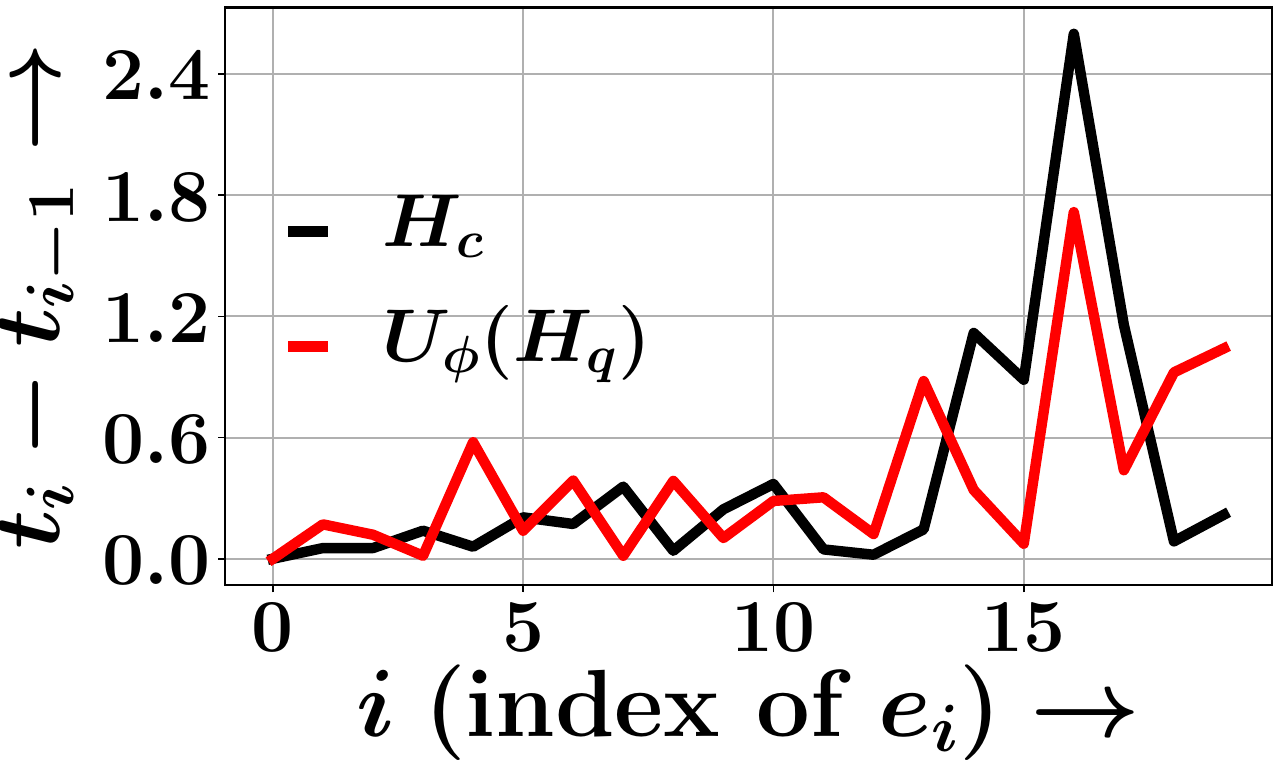}
  \caption{$U_{\phi}(\Hcal_q), \Hcal_c$}
\end{subfigure}
\caption{Effect of unwarping on a \emph{relevant} query-corpus pair in Audio dataset. $U_{\phi}(\cdot)$ learns to transform $\Hcal_q$ in order to capture a high value of its latent similarity with $\Hcal_c$.}
\label{fig:UU}
\end{figure}

In all cases, we used \nsrs. Table~\ref{tab:model_ablation} shows that the complete design of our model~(variant \eqref{var:our})  achieves the best performance. We further note that removing $\kernel_{p_\theta}$ from the score (variant \eqref{var:withoutmodel}) leads to significantly poor performance. Interestingly, our model without any mark-based similarity~(variant \eqref{var:withoutmark}) leads to better performance than the model without time similarity  (variant~\eqref{var:withouttime})--- this could be attributed to the larger variance in query-corpus time distribution than the distribution of marks. 

\xhdr{Effect of $U_{\phi}(\cdot)$}
Finally, we observe that the performance deteriorates if we do not use an unwarping function $U_{\phi}(\cdot)$ (variant~\eqref{var:withoutU}). Figure~\ref{fig:UU} illustrates the effect of $U_{\phi}(\cdot)$. It shows that $U_{\phi}(\cdot)$ is able to learn a suitable transformation of the query sequence, which encapsulates the high value of latent similarity with the corpus sequence.

\begin{figure}[t]
\centering
\begin{subfigure}{0.45\columnwidth}
  \centering
  \includegraphics[height=3cm]{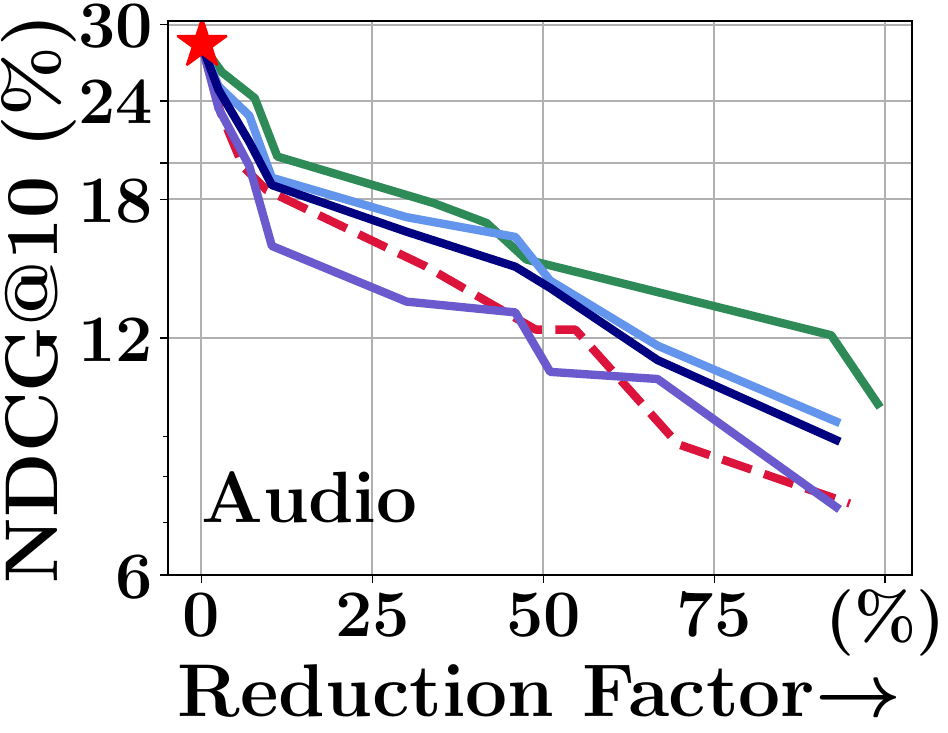}
  \caption{Audio}
\end{subfigure}
\hspace{0.5cm}
\begin{subfigure}{0.45\columnwidth}
  \centering
  \includegraphics[height=3cm]{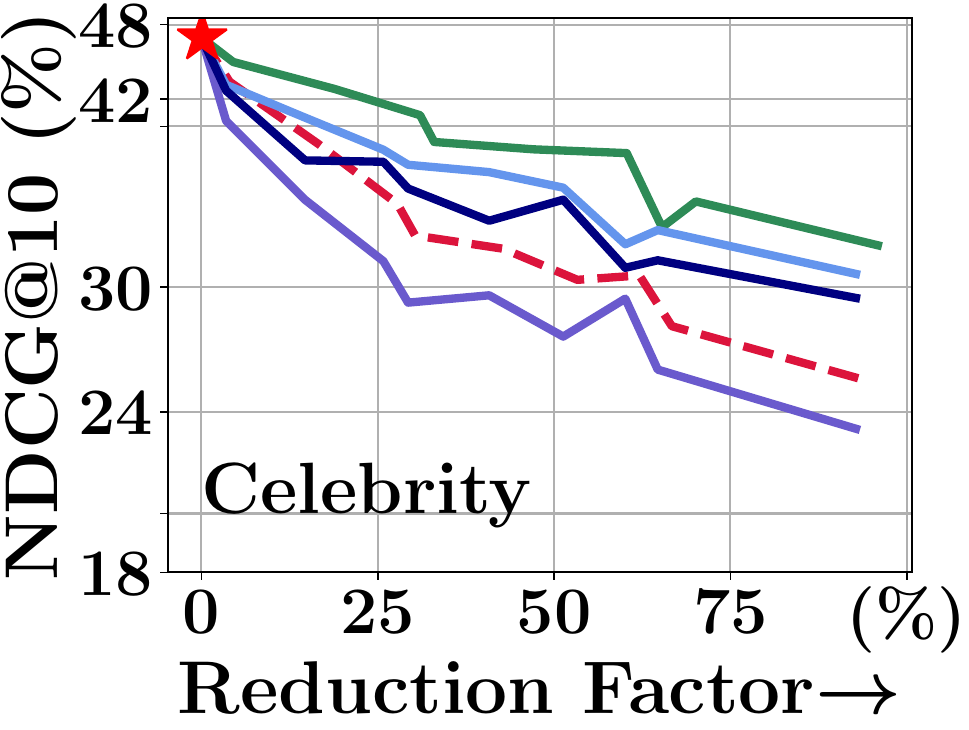}
  \caption{Celebrity}
\end{subfigure}

\centering
\begin{subfigure}{0.45\columnwidth}
  \centering
  \includegraphics[height=3cm]{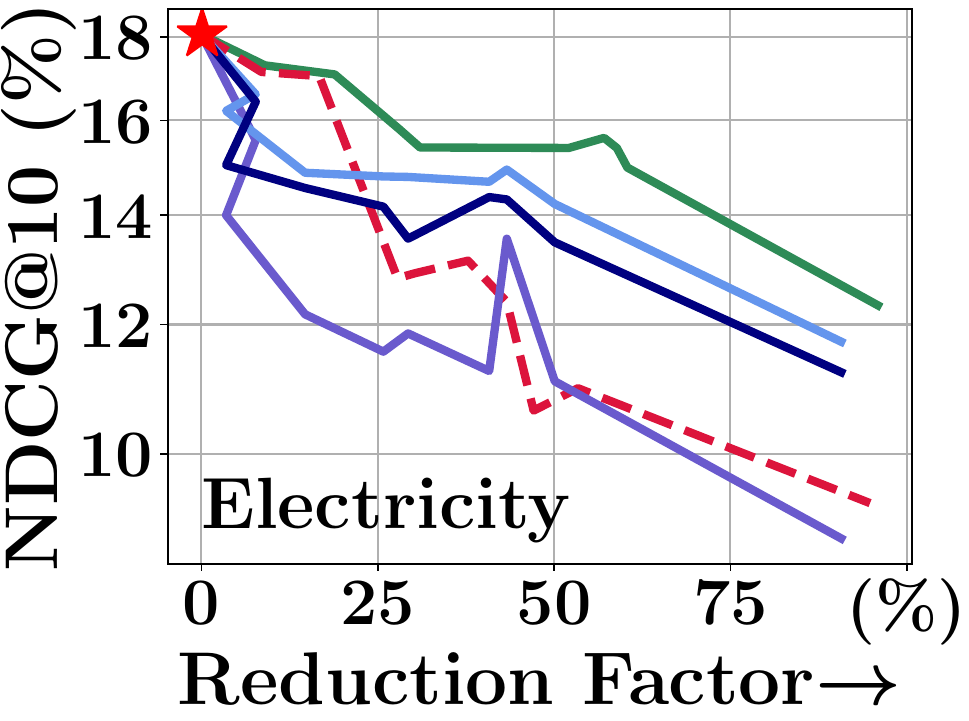}
  \caption{Electricity}
\end{subfigure}
\hspace{0.5cm}
\begin{subfigure}{0.45\columnwidth}
  \centering
  \includegraphics[height=3cm]{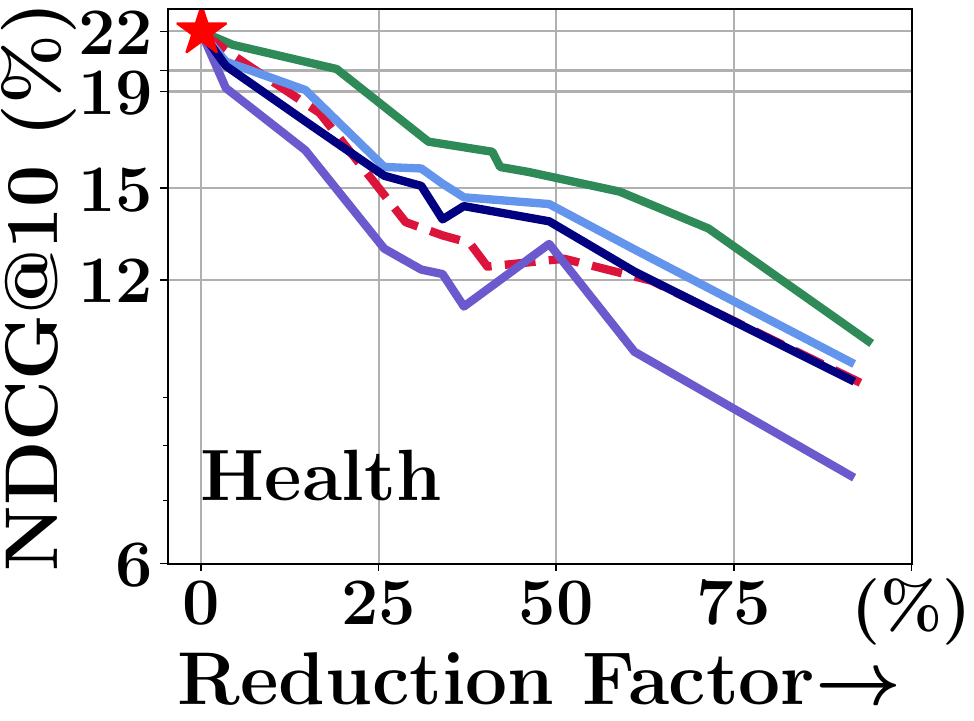}
  \caption{Health}
\end{subfigure}
{\includegraphics[height=1cm]{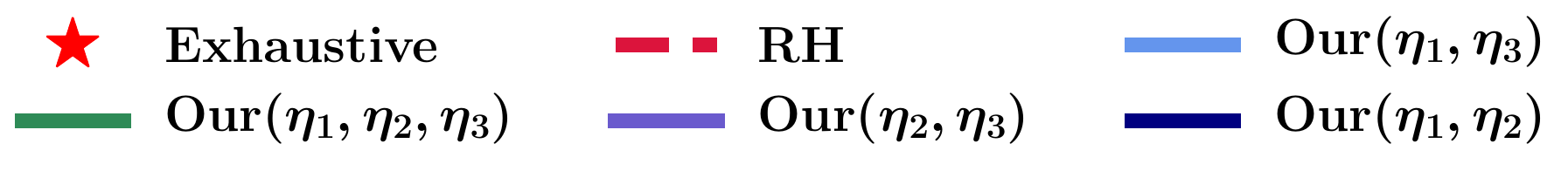}}
\caption{Tradeoff between NDCG@10 vs. Reduction factor, \ie, \% reduction in number of comparisons between query-corpus pairs w.r.t. the exhaustive comparisons for different hashing methods. The point marked as $ \color{red} {\star}$ indicates the case with exhaustive comparisons on the set of corpus sequences. }
\label{fig:hash}
\end{figure}

\subsection{Results on Retrieval Efficiency}
We compare our efficient sequence retrieval method  given in Algorithm~\ref{alg:key} against random hyperplane (RH) method and three variants of our proposed training problem in Eq.~\eqref{eq:hash}. 
\begin{itemize}
\item Our $(\eta_2, \eta_3)$ which sets $\eta_1=0$ and thus does not enforce even distribution of $\pm 1 $ in $\hash^c$;
\item Our $(\eta_1, \eta_3)$ which sets $\eta_2=0$ and thus $\tanh$ does not accurately approximate $\sgn$;
\item Our $(\eta_1, \eta_2)$ which sets $\eta_3=0$ and thus does not enforce $\hash^c$ to be compact and free of redundancy. 
\end{itemize}
Our $(\eta_1,\eta_2,\eta_3)$ is the complete design which includes all trainable components. Figure~\ref{fig:hash} summarizes the results.

\xhdr{Comparison with random hyperplane}   Figure~\ref{fig:hash}  shows that our method (Our$(\eta_1,\eta_2,\eta_3)$) demonstrates better Pareto efficiency than RH. This is because RH generates hash code in a data oblivious manner whereas our method learns the hash code on top of the trained embeddings.

\xhdr{Ablation Study on Different Components of Eq.~\eqref{eq:hash}}
Figure~\ref{fig:hash} summarizes the results, which shows that (i) the first three variants are outperformed by Our$(\eta_1,\eta_2,\eta_3)$; (ii) the first term having $\eta_1 \neq 0$, which enforces an even distribution of $\pm 1$, is the most crucial component for the loss function--- as the removal of this term causes significant deterioration of the performance.

\subsection{Analysis at a Query Level} 
\begin{figure}[t]
\centering
\begin{subfigure}{0.45\columnwidth}
  \centering
  \includegraphics[height=3cm]{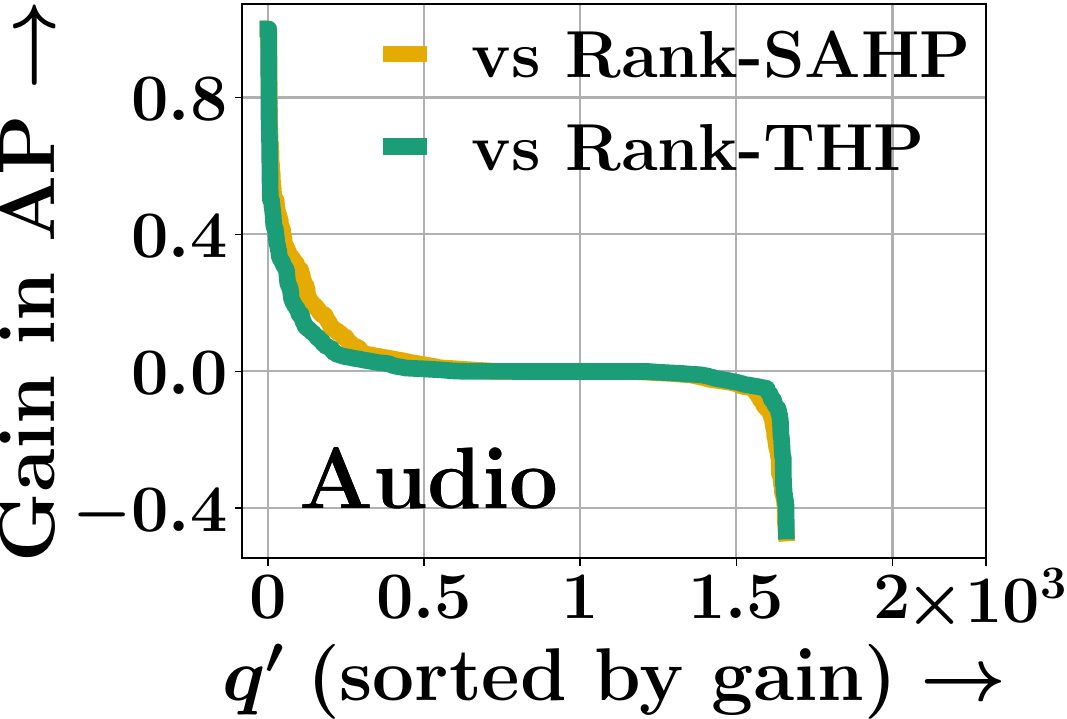}
  \caption{Audio}
\end{subfigure}
\hspace{0.5cm}
\begin{subfigure}{0.45\columnwidth}
  \centering
  \includegraphics[height=3cm]{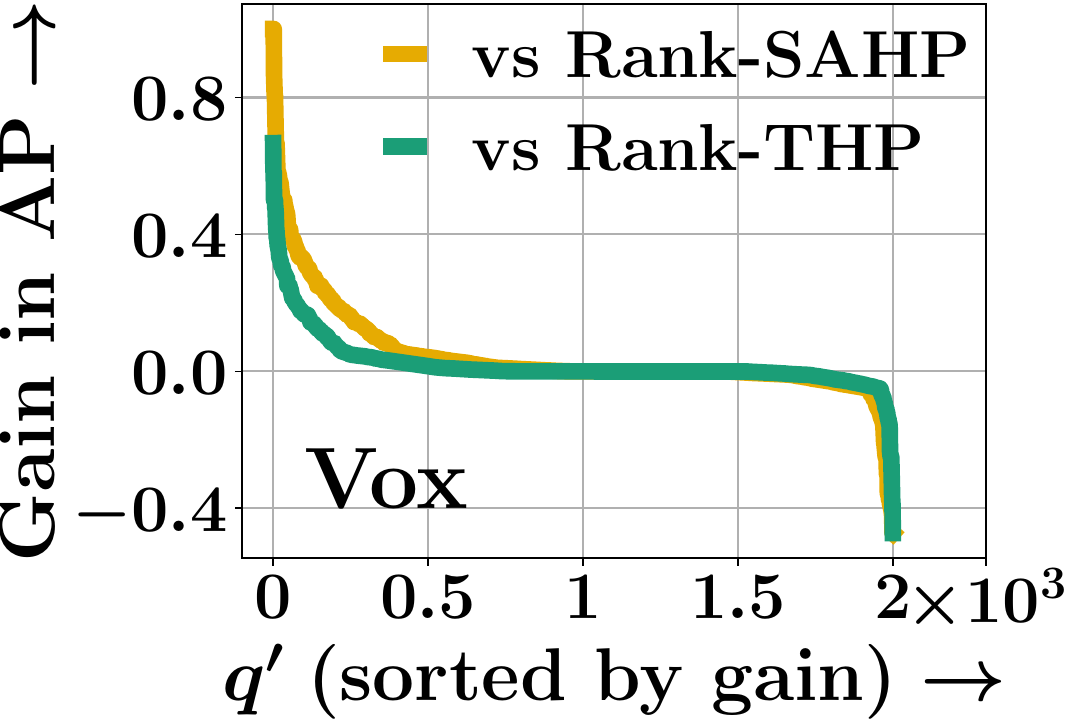}
  \caption{Celebrity}
\end{subfigure}

\centering
\begin{subfigure}{0.45\columnwidth}
  \centering
  \includegraphics[height=3cm]{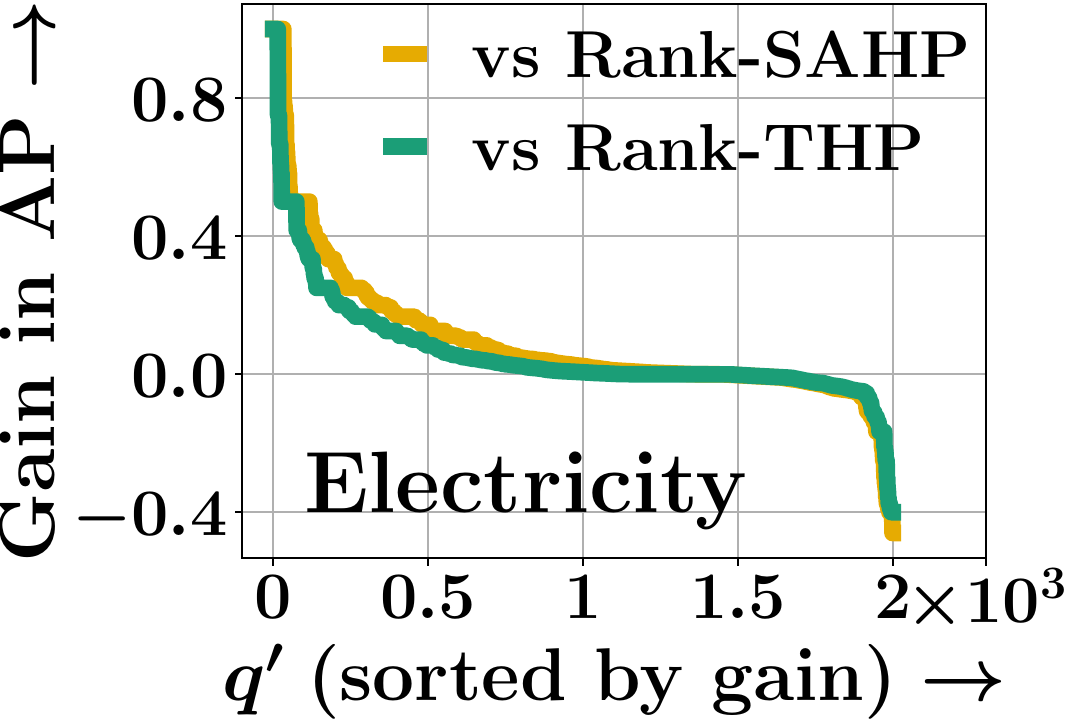}
  \caption{Electricity}
\end{subfigure}
\hspace{0.5cm}
\begin{subfigure}{0.45\columnwidth}
  \centering
  \includegraphics[height=3cm]{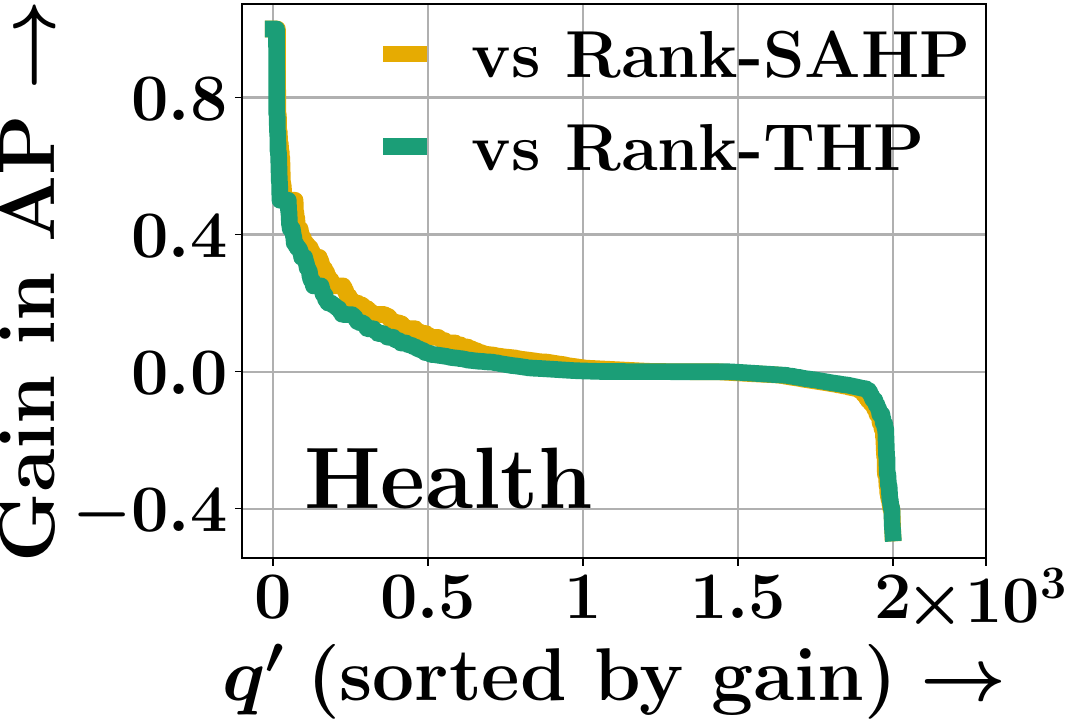}
  \caption{Health}
\end{subfigure}
\caption{Query-wise performance comparison between \nsr and best baseline methods -- \tpprank-THP, \tpprank-SAHP. Queries are sorted by the decreasing gain in AP.}
\label{fig:app_drill}
\end{figure}

Next, we compare the performance between \nsr and other state-of-the-art methods, at a query level.  Specifically, for each query $\Hcal_q$ we compute the advantage of using \nsr in terms of gain in average precision, \ie, AP(\nsr) $-$ AP(baseline) for two most competitive baselines -- \tpprank-SAHP and \tpprank-THP.  We summarize the results in Figure~\ref{fig:app_drill}, which show that for at least 70\% of the queries, \nsr outperforms or fares competitively with these baselines, across all datasets.

\begin{table}[t!]
    \centering
    \caption{Training-times of \nsr for all datasets.}
    \resizebox{\textwidth}{!}{%
    \begin{tabular}{c|ccccccc}
    \toprule
    \textbf{Run Time} & \textbf{Audio} & \textbf{Celebrity} & \textbf{Electricity} & \textbf{Health} & \textbf{Sports} & \textbf{Twitter} & \textbf{Amazon}\\ \midrule
    \nsr & $\le$ 3hr & $\le$ 5hr & $\le$ 6hr & $\le$ 6hr & $\le$ 2hr & $\le$ 3hr & $\le$ 3hr\\
    \nsrp & $\le$ 4hr & $\le$ 5hr & $\le$ 6hr & $\le$ 6hr & $\le$ 3hr & $\le$ 3hr & $\le$ 3hr\\ 
    \midrule
    \end{tabular}
    }
    \label{tab:runtime}
\end{table}

\begin{table}[t!]
    \caption{Retrieval quality in terms of mean average precision (MAP) for query sequence lengths sampled between 10 and 50.}
    \centering
    \resizebox{0.9\textwidth}{!}{%
    \begin{tabular}{l|ccccccc}
    \hline
    \textbf{$|\mathcal{H}_q| = (10,20)$} & \textbf{Audio} & \textbf{Celebrity} & \textbf{Health} & \textbf{Electricity} & \textbf{Sports} & \textbf{Twitter} & \textbf{Amazon}\\ \hline
    \texttt{Rank}-RMTPP & 17.83 & 21.46 & 9.23 & 11.76 & 25.39 & 18.76 & 26.38\\
    \texttt{Rank}-THP & 18.97 & 22.06 & 9.48 & 12.27 & 26.58 & 19.49 & 27.61\\
    \nsrs & 21.30 & 25.77 & 15.83 & 10.79 & \textbf{27.63} & 22.63 & 28.71\\
    \nsrp & \textbf{21.87} & \textbf{26.14} & \textbf{16.42} & \textbf{11.63} & 27.18 & \textbf{22.86} & \textbf{28.97}\\
    \hline
    \end{tabular}
    }
    \label{tab:length1}
    \centering
    \caption{Retrieval quality in terms of mean average precision (MAP) for query sequence lengths sampled between 50 and 100.}
    \resizebox{0.9\textwidth}{!}{%
    \begin{tabular}{l|ccccccc}
    \hline
    \textbf{$|\mathcal{H}_q| = (50,100)$} & \textbf{Audio} & \textbf{Celebrity} & \textbf{Health} & \textbf{Electricity} & \textbf{Sports} & \textbf{Twitter} & \textbf{Amazon}\\ \hline
    \texttt{Rank}-RMTPP & 24.31 & 33.78 & 15.62 & 20.84 & 30.73 & 37.38 & 42.69\\
    \texttt{Rank}-THP & 27.94 & 37.85 & 17.58 & 21.61 & 31.26 & 38.94 & 43.98\\
    \nsrs & 28.58 & 41.92 & 19.67 & 23.54 & 36.92 & 45.65 & \textbf{49.83}\\
    \nsrp & \textbf{29.73} & \textbf{42.84} & \textbf{21.48} & \textbf{24.81} & \textbf{38.03} & \textbf{46.71} & 49.47\\
    \hline
    \end{tabular}
    }
    \label{tab:length2}
\end{table}

\subsection{Runtime Analysis}
Next, we calculate the run-time performance of \nsr. With this experiment, our goal is to determine if the training times of \nsr are suitable for designing solutions for real-world problems. From the results in Table~\ref{tab:runtime}, we note that even for datasets with up to 60 million events, the training times are well within the feasible range for practical deployment.

\subsection{Query Length}
We perform an additional experiment of sequence retrieval with varying query lengths. Specifically, we sample queries of lengths $|\mathcal{H}_q| \sim Unif(10, 50)$ and $|\mathcal{H}_q| \sim Unif(50, 100)$ and report the sequence retrieval results in Table~\ref{tab:length1} and Table~\ref{tab:length2} respectively. The results show that the performance of all models deteriorates significantly as we reduce the length of query sequences. They also show that even with smaller query lengths, \nsrs significantly outperforms the other state-of-the-art baseline \texttt{Rank}-THP. Moreover, here we also note that the performance gains due to \nsrp are more significant than with complete sequences. This further supports the need to use a cross-attention mechanism to capture the dynamics between sequences. 

\subsection{Qualitative Analysis}
\begin{figure}
\centering
\begin{subfigure}{\columnwidth}
  \centering
 {\includegraphics[height=2.5cm]{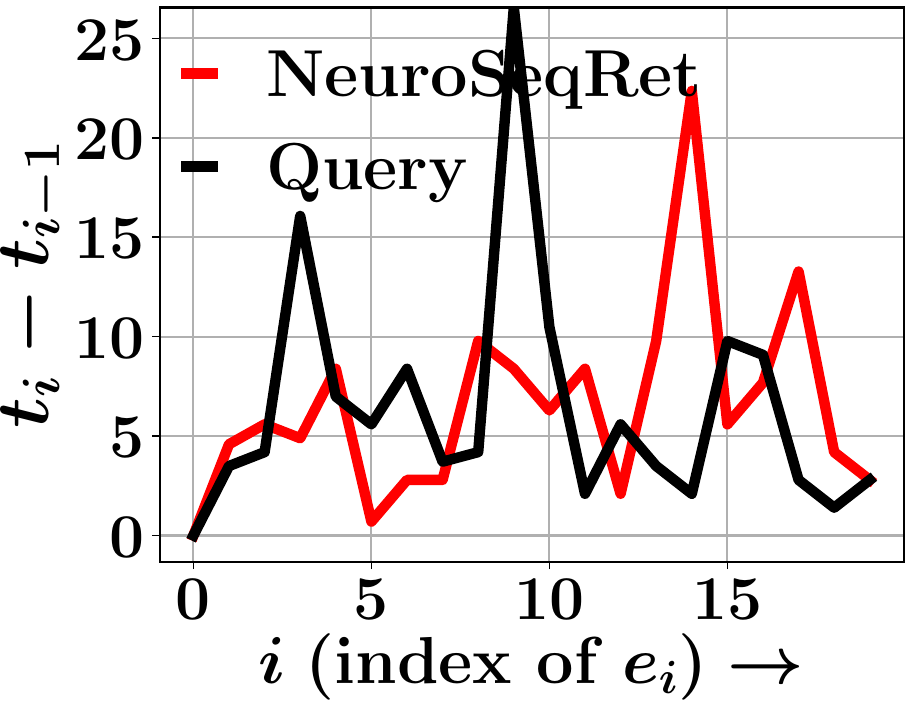}}
 {\includegraphics[height=2.5cm]{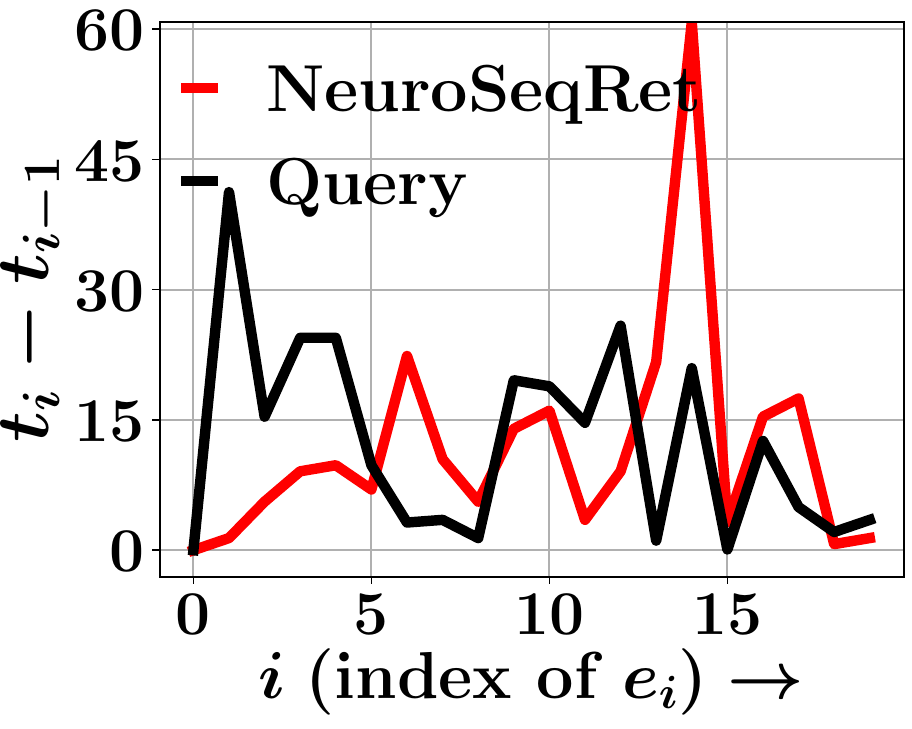}}
 {\includegraphics[height=2.5cm]{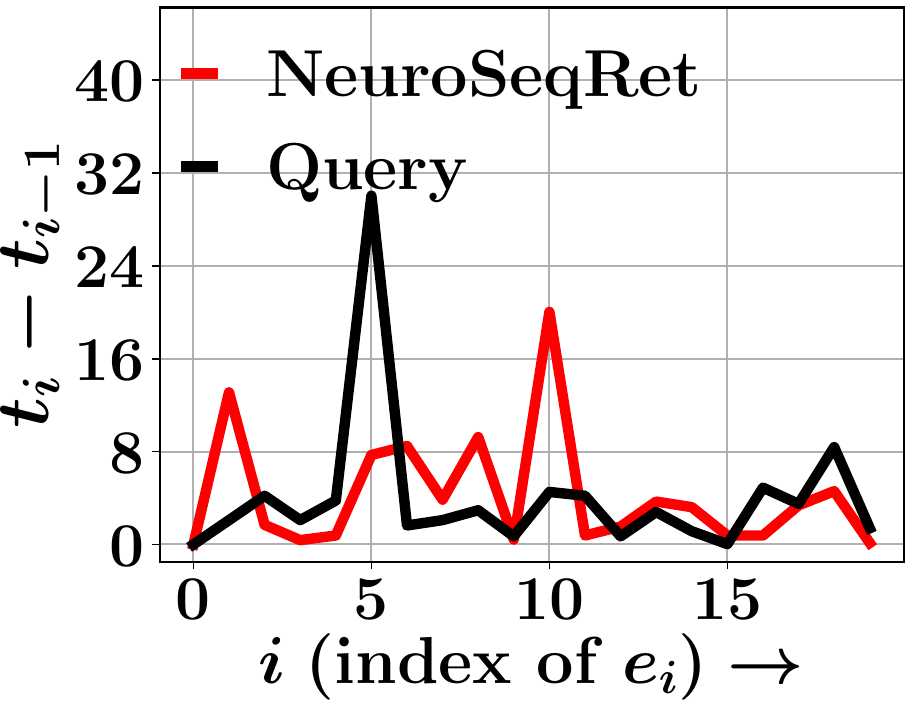}}
 {\includegraphics[height=2.5cm]{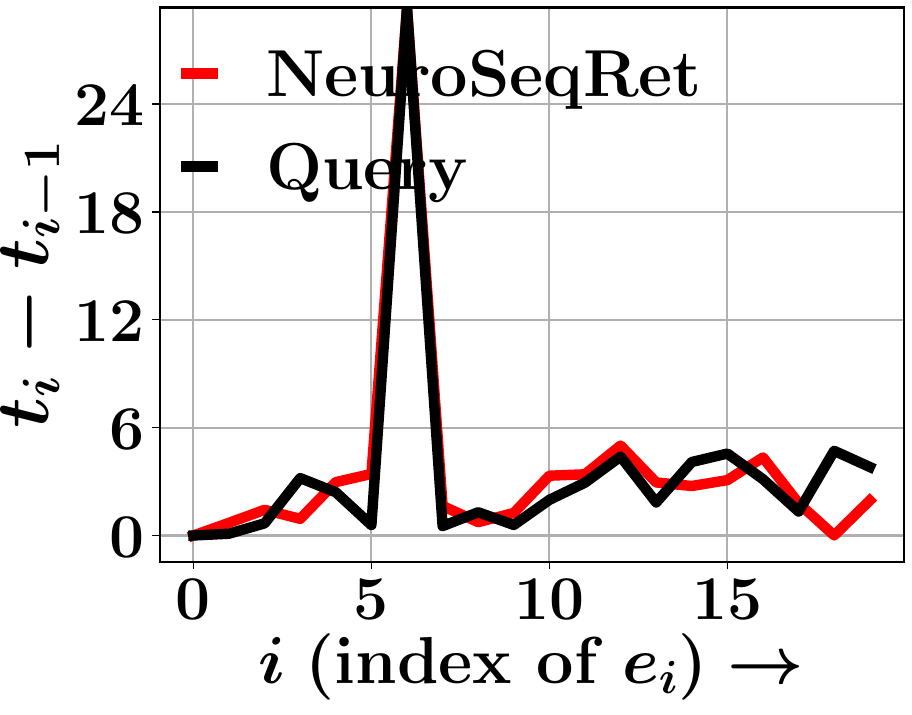}}
  \caption{Audio}
\end{subfigure}

\begin{subfigure}{\columnwidth}
  \centering
{\includegraphics[height=2.5cm]{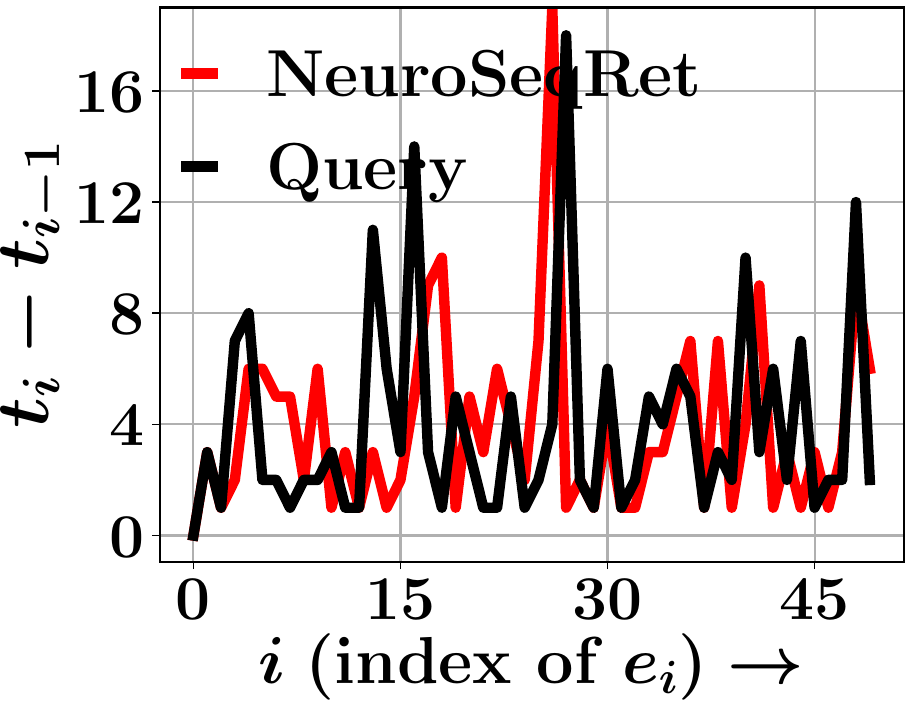}}
{\includegraphics[height=2.5cm]{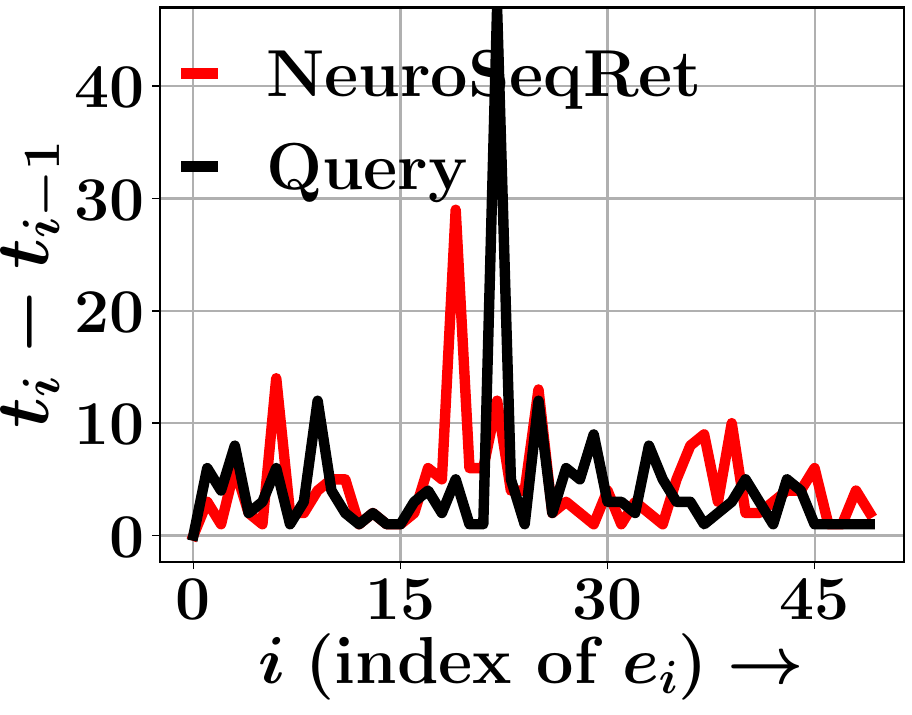}}
{\includegraphics[height=2.5cm]{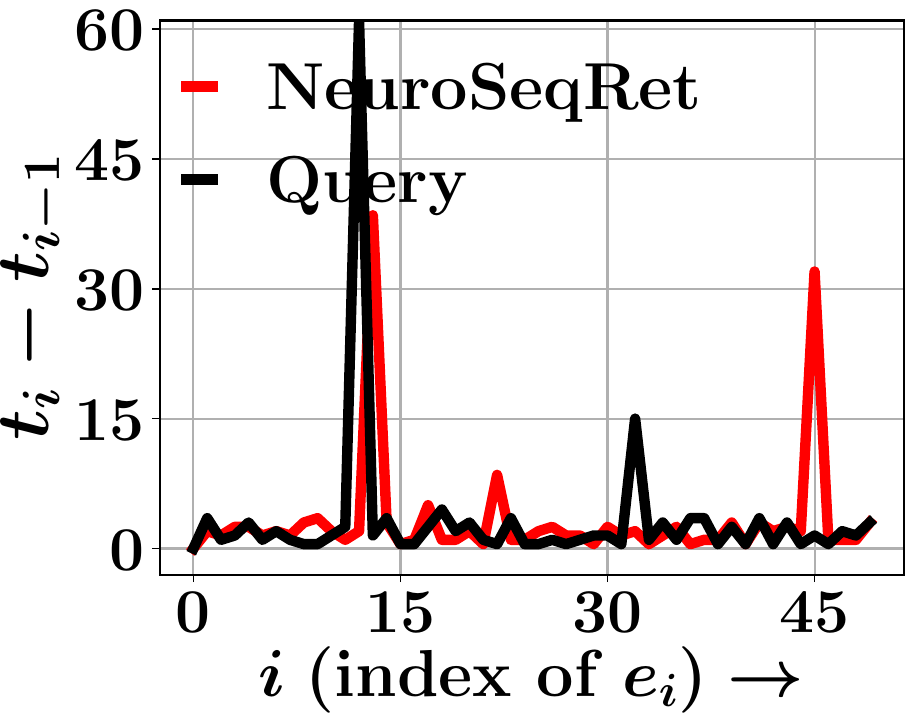}}
{\includegraphics[height=2.5cm]{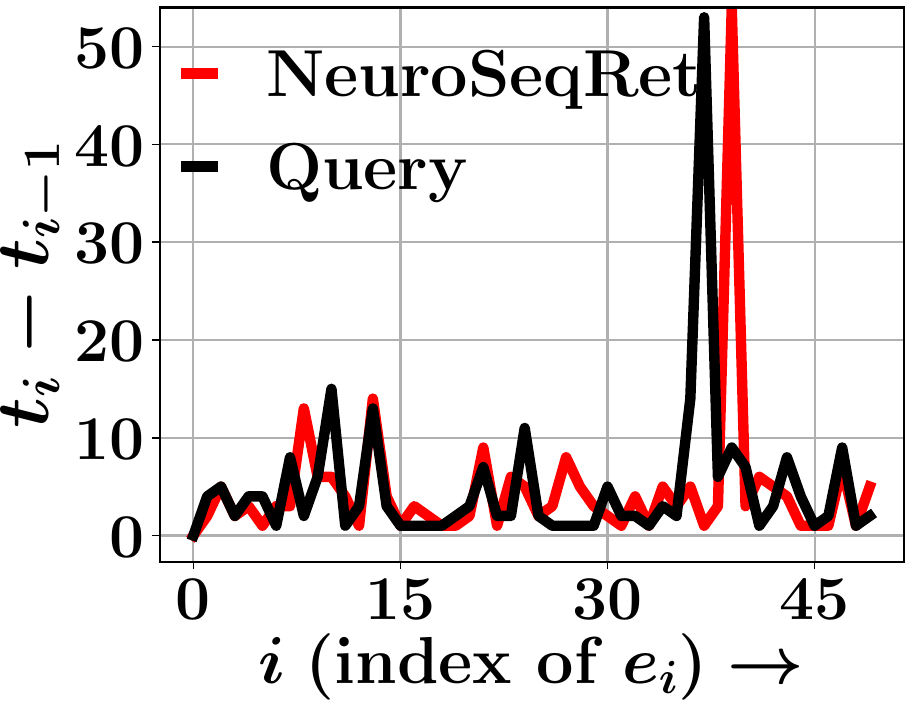}}
  \caption{Celebrity}
\end{subfigure}

\begin{subfigure}{\columnwidth}
  \centering
{\includegraphics[height=2.5cm]{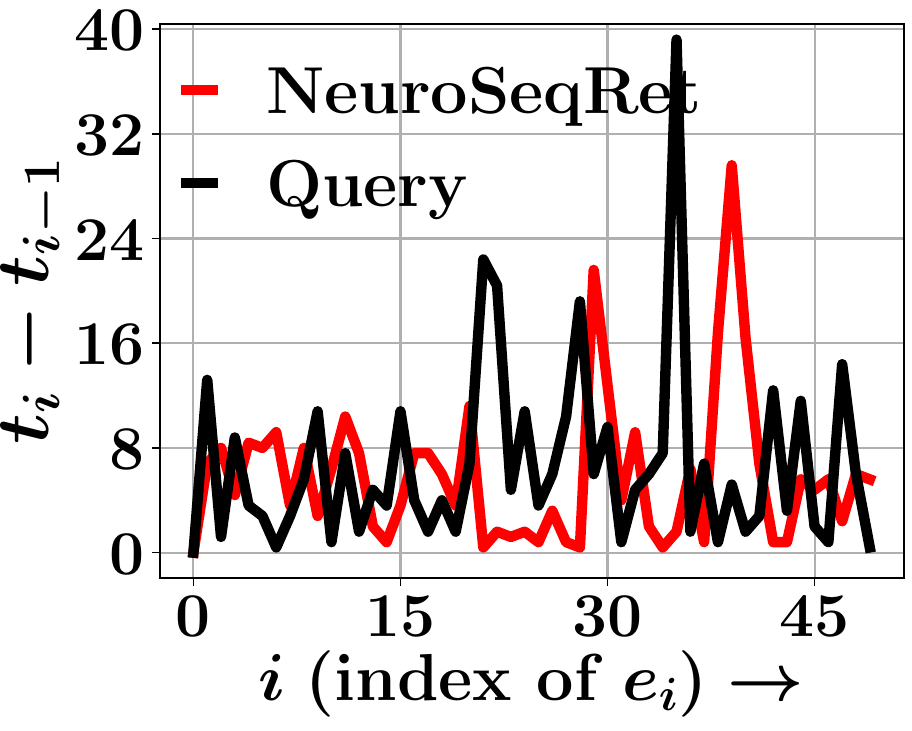}}
{\includegraphics[height=2.5cm]{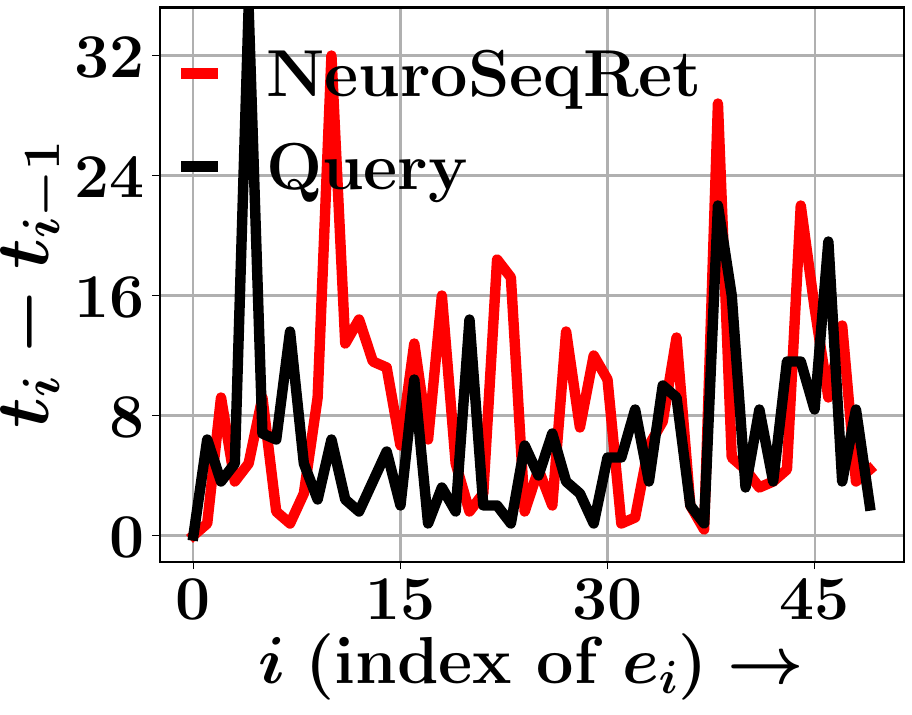}}
{\includegraphics[height=2.5cm]{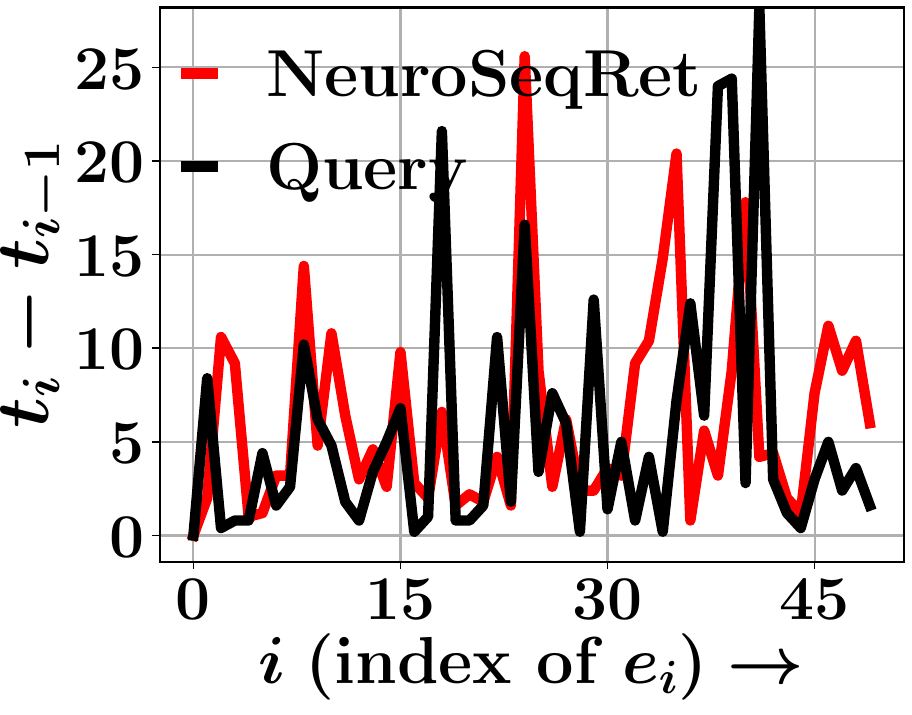}}
{\includegraphics[height=2.5cm]{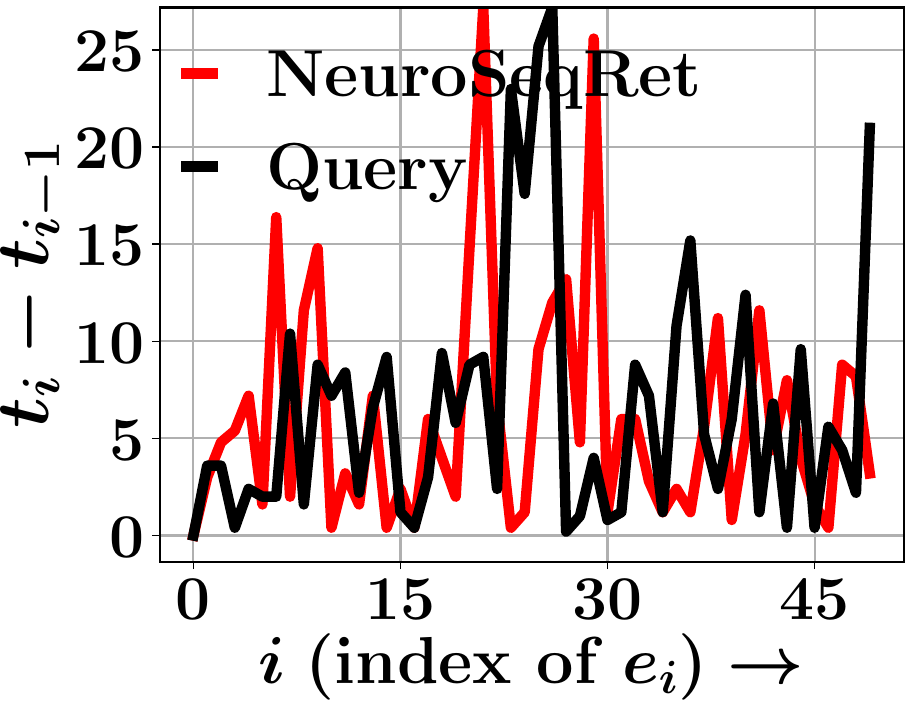}}
  \caption{Electricity}
\end{subfigure}

\begin{subfigure}{\columnwidth}
  \centering
{\includegraphics[height=2.5cm]{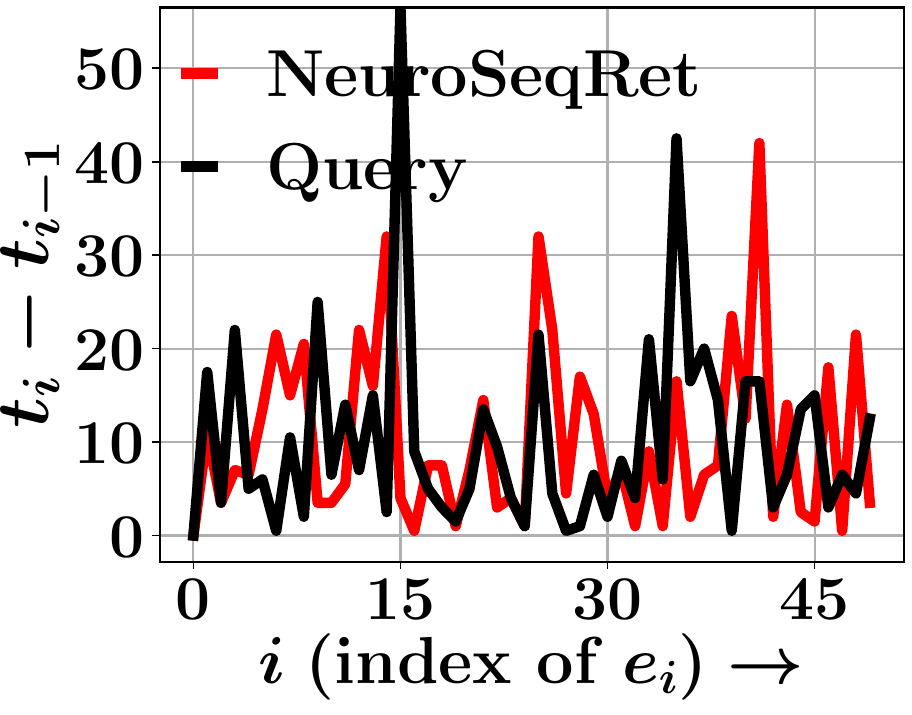}}
{\includegraphics[height=2.5cm]{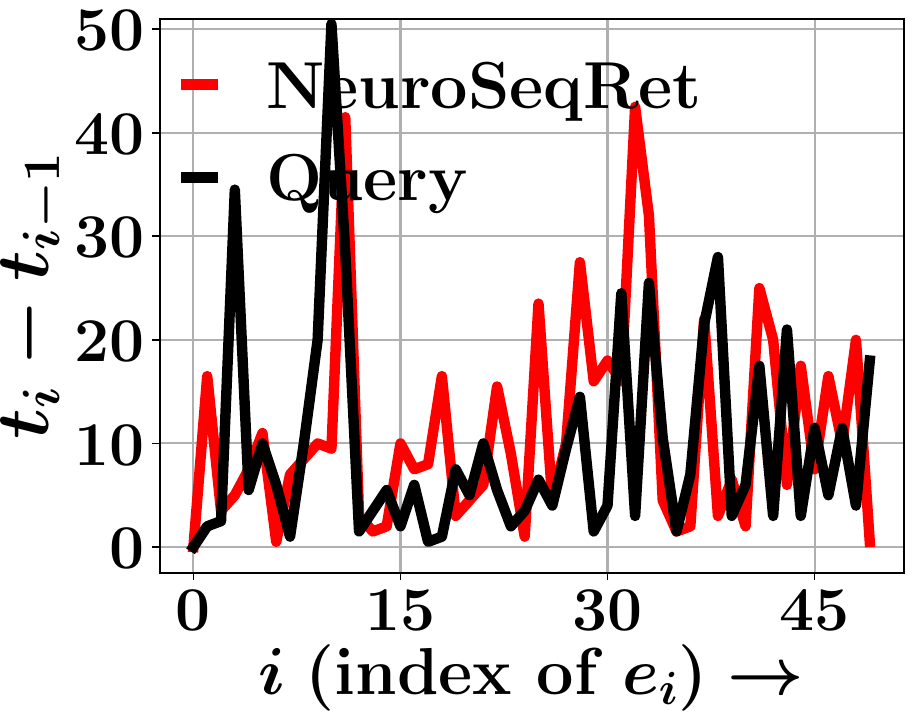}}
{\includegraphics[height=2.5cm]{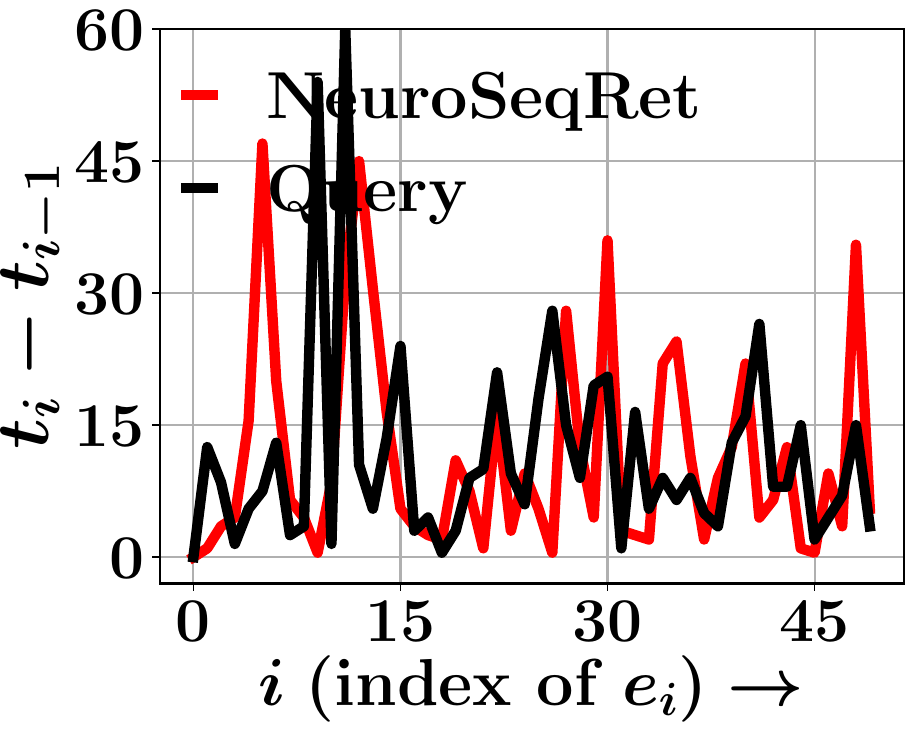}}
{\includegraphics[height=2.5cm]{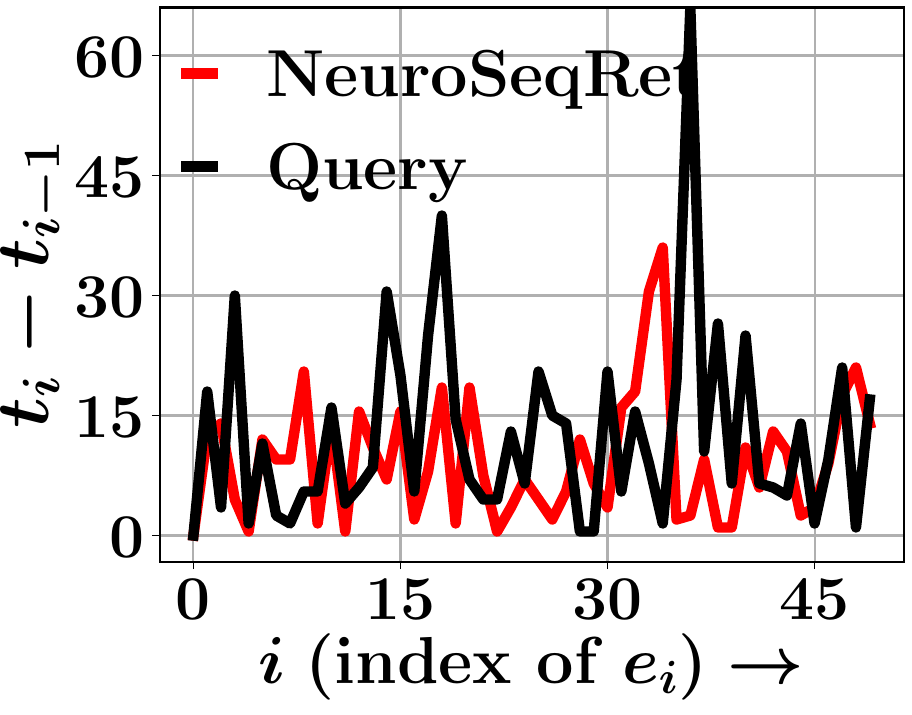}}
  \caption{Health}
\end{subfigure}
\caption{Qualitative examples of inter-event times of events in a query sequence and the top-search results by \nsr for four datasets.}
\label{fig:qualitative}
\end{figure}

To get deeper insights into the working of our model, we perform a qualitative analysis between a query sequence from the dataset and the sequence retrieved by \nsr. More specifically, we aim to understand the similar patterns between query and corpus sequences that \nsr searches for in the corpus and plot the query sequence and the corresponding top-ranked relevant corpus sequence retrieved by \nsr. The results across four datasets -- Audio, Celebrity, Electricity, and Health -- in Figure~\ref{fig:qualitative} show that the inter-arrival times of the CTES retrieved by \nsr closely match the query inter-arrival times. For brevity, we have excluded the plots for Sports, Amazon, and Twitter, however, we noted that the plots for these datasets also showed a similar trend. 

\section{Conclusion}  \label{sec:conc} 
In this paper, we proposed a novel supervised continuous-time event sequence retrieval system called \nsr\ using neural MTPP models.  To achieve efficient retrieval over a very large corpus of sequences, we also propose a trainable hash-coding of corpus sequences which can be used to narrow down the number of sequences to be considered for similarity score computation. Our experiments with real-world datasets from a diverse range of domains show that our retrieval model is more effective than several baselines. Our work opens several avenues for future work including the design of generative models for relevance sequences and counter-factual explanations for relevance label predictions. Currently, in high-impact applications like healthcare, the use of our method requires additional care, since an incorrect prediction made by our model would have an adverse impact. In the end, one may consider a human-in-loop retrieval system, that will mitigate such risk via human intervention. Moreover, event sequences from some domains, \eg, mobility records, can contain user-specific data and their use may lead to privacy violations. To that aim, it would be interesting to consider designing a privacy-preserving retrieval system for continuous-time event sequences. Empirically, we show that our hash-code-based retrieval is Pareto-efficient, offering a trade-off between the computational cost and the retrieval performance.

\bibliographystyle{ACM-Reference-Format}
\bibliography{refs}
\end{document}